\documentclass{article}

\usepackage{microtype}
\usepackage{graphicx}
\usepackage{subfigure}
\usepackage{booktabs} % for professional tables
\usepackage{kotex}
\usepackage{diagbox}
\usepackage{tabularx}
\usepackage{multirow}
\usepackage{booktabs}

\usepackage{makecell}
\usepackage{hyperref}

\usepackage[accepted]{icml2025}

\usepackage{amsmath}
\usepackage{amssymb}
\usepackage{mathtools}
\usepackage{amsthm}

\usepackage[capitalize,noabbrev]{cleveref}

\theoremstyle{plain}

\theoremstyle{definition}

\theoremstyle{remark}

\usepackage{bbm}

\usepackage[textsize=tiny]{todonotes}

\icmltitlerunning{Submission and Formatting Instructions for ICML 2025}

\begin{document}

\twocolumn[
\icmltitle{Domain-Invariant Per-Frame Feature Extraction for Cross-Domain Imitation Learning with Visual Observations}

% It is OKAY to include author information, even for blind
% submissions: the style file will automatically remove it for you
% unless you've provided the [accepted] option to the icml2025
% package.

% List of affiliations: The first argument should be a (short)
% identifier you will use later to specify author affiliations
% Academic affiliations should list Department, University, City, Region, Country
% Industry affiliations should list Company, City, Region, Country

% You can specify symbols, otherwise they are numbered in order.
% Ideally, you should not use this facility. Affiliations will be numbered
% in order of appearance and this is the preferred way.
\icmlsetsymbol{equal}{*}

\begin{icmlauthorlist}
\icmlauthor{Minung Kim}{uni}
\icmlauthor{Kawon Lee}{uni}
\icmlauthor{Sungho Choi}{kai}
\icmlauthor{Jungmo Kim}{uni}
\icmlauthor{Seungyul Han}{uni} \hspace{-0.12in} $^{*}$
\end{icmlauthorlist}

\icmlaffiliation{uni}{Artificial Intelligence Graduate School, UNIST, Ulsan, South Korea}
\icmlaffiliation{kai}{School of Electrical Engineering, KAIST, Daejeon, South Korea}

% \icmlcorrespondingauthor{Minung Kim}{minungkim@unist.ac.kr}
\icmlcorrespondingauthor{Seungyul Han}{syhan@unist.ac.kr}

% You may provide any keywords that you
% find helpful for describing your paper; these are used to populate
% the "keywords" metadata in the PDF but will not be shown in the document
\icmlkeywords{Machine Learning, ICML}

\vskip 0.3in
]

% this must go after the closing bracket ] following \twocolumn[ ...

% This command actually creates the footnote in the first column
% listing the affiliations and the copyright notice.
% The command takes one argument, which is text to display at the start of the footnote.
% The \icmlEqualContribution command is standard text for equal contribution.
% Remove it (just {}) if you do not need this facility.

\printAffiliationsAndNotice{}  % leave blank if no need to mention equal contribution
% \printAffiliationsAndNotice{\icmlEqualContribution} % otherwise use the standard text.

\begin{abstract}
% This document provides a basic paper template and submission guidelines.
% Abstracts must be a single paragraph, ideally between 4--6 sentences long.
% Gross violations will trigger corrections at the camera-ready phase.
Imitation learning (IL) enables agents to mimic expert behavior without reward signals but faces challenges in cross-domain scenarios with high-dimensional, noisy, and incomplete visual observations. To address this, we propose Domain-Invariant Per-Frame Feature Extraction for Imitation Learning (DIFF-IL), a novel IL method that extracts domain-invariant features from individual frames and adapts them into sequences to isolate and replicate expert behaviors. We also introduce a frame-wise time labeling technique to segment expert behaviors by timesteps and assign rewards aligned with temporal contexts, enhancing task performance. Experiments across diverse visual environments demonstrate the effectiveness of DIFF-IL in addressing complex visual tasks.

\end{abstract}
%==========================================================

\section{Introduction}
\label{sec:intro}
%===================

Imitation learning (IL) allows agents to learn complex behaviors by observing and replicating expert demonstrations without requiring explicit reward signals. This approach is widely applied in robotics, autonomous driving, and healthcare. The simplest IL technique, behavior cloning (BC) \cite{bain1995framework, pomerleau1991efficient, Ross2011Reduction, Torabi2018behavioral}, directly mimics expert datasets but struggles with generalization when the agent deviates from training trajectories. Inverse reinforcement learning (IRL) addresses this limitation by inferring reward functions from expert behavior, enabling more robust learning \cite{Ng2000, Abbeel2004, Ziebart2008}. Adversarial imitation learning (AIL) builds on IRL by aligning state-action distributions between learners and experts using adversarial frameworks \cite{Finn2016, Fu2018, Ho2016, Torabi2018generative, Zhang2020}, often utilizing generative models like Generative Adversarial Networks (GANs) \cite{Goodfellow2014}. While effective in same-domain scenarios, these methods face challenges in cross-domain settings due to domain shifts that complicate policy transfer \cite{ben2006analysis}.

In cross-domain scenarios, mismatches arise from differences in viewpoints, dynamics, embodiments, and state spaces, creating significant hurdles for IL applications. For instance, autonomous driving may require learning from simulations while operating in real-world environments, or robots may rely on visual data to control physical joints. These shifts exacerbate learning difficulties, particularly with high-dimensional and noisy visual data, where even minor variations can disrupt alignment and stability. To address these issues, cross-domain IL techniques extract domain-invariant features from visual datasets to align source and target domains while retaining task-relevant information \cite{li2018oil, Liu2018, Cetin2021, shang2021, choi2024domain}. By focusing on features independent of domain-specific factors, these methods enable learners to mimic expert behavior using visual demonstrations, improving IL's effectiveness across diverse real-world scenarios \cite{sermanet2018}.

\begin{figure*}[ht]
    \includegraphics[width=\textwidth]{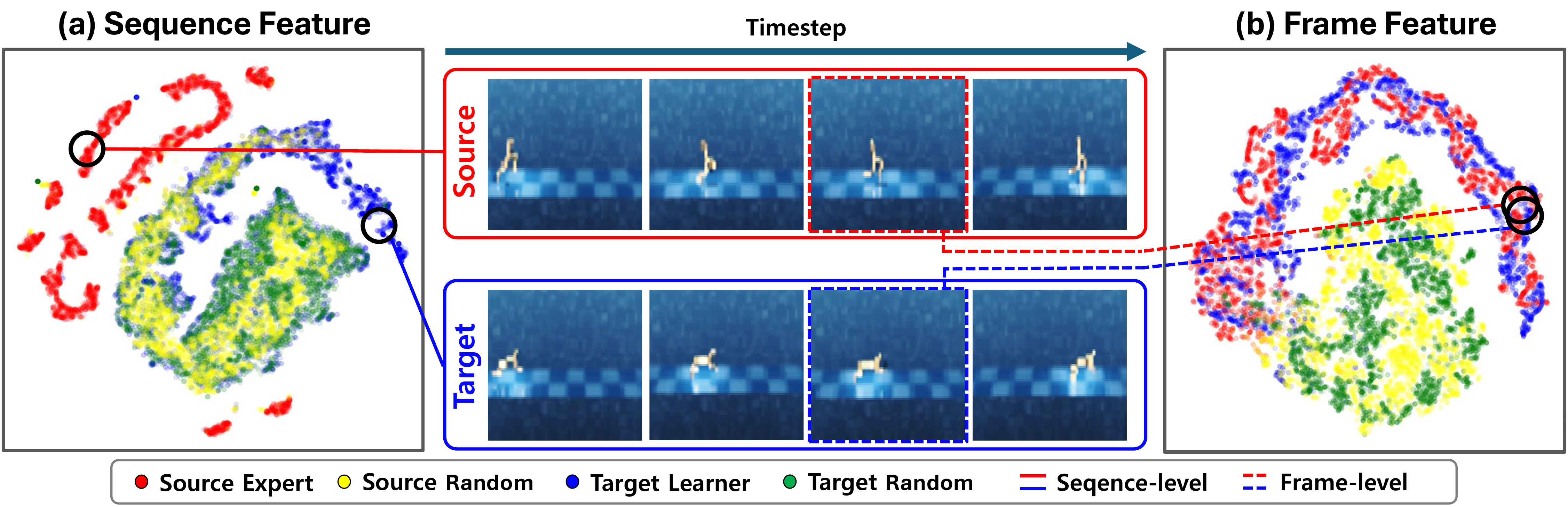}
    \vspace{-0.2in}
    \caption{t-SNE visualization of features extracted from:
(a) Existing sequence-based IL methods, (b) DIFF-IL (ours).}
    \vspace{-0.1in}
    \label{fig:motiv}
\end{figure*}

Existing IL methods often rely on image sequences spanning multiple timesteps to identify domain-invariant features for IRL and reward design, as single images cannot fully capture an agent's evolving behavior. However, these approaches frequently struggle with the complexity of sequence spaces, leading to misaligned features, poorly designed rewards, and suboptimal imitation of expert policies. To address these challenges, we propose Domain-Invariant Per-Frame Feature Extraction for Imitation Learning (DIFF-IL). DIFF-IL introduces two key contributions: (1) per-frame domain-invariant feature extraction to robustly isolate domain-independent task-relevant behaviors, and (2) frame-wise time labeling to segment expert behaviors by timesteps and assign rewards based on temporal alignment. Together, these innovations enable precise domain alignment and effective imitation, even in scenarios with limited overlap between source domain data and expert actions.

Fig. \ref{fig:motiv} highlights the strengths of the proposed DIFF-IL method in the Walker (source)-to-Cheetah (target) environment. In this scenario, the Cheetah agent (target learner) aims to move forward quickly by mimicking expert demonstrations from Walker agents (source expert), despite differing dynamics. Fig. \ref{fig:motiv}(a) presents a t-SNE visualization of latent features from sequences of four consecutive image frames extracted using sequence-based IL methods. While source expert and target learner agents share similar positions and speeds, their features fail to align due to residual domain-specific information, leading to inaccurate rewards and suboptimal learning. In contrast, Fig. \ref{fig:motiv}(b) shows that DIFF-IL seamlessly aligns latent features of individual image frames across domains, effectively removing domain-specific artifacts while preserving expertise-critical details. This enables DIFF-IL to extract truly domain-invariant features, allowing the learner to accurately mimic expert behaviors, unlike sequence-based methods that fail to achieve robust domain adaptation. Moreover, in scenarios where source expert and random behaviors overlap minimally, traditional methods often misclassify slight deviations from random as expert behavior, hindering learning. DIFF-IL addresses this by incorporating frame-wise time labeling, which segments expert behaviors into finer temporal contexts. By assigning higher rewards to frames closer to the goal, DIFF-IL guides the agent to progressively replicate expert trajectories, ensuring robust alignment and successful task completion, even under challenging conditions.

\section{Related Works}
\label{sec:related}

\vspace{-0.1in}
\textbf{Imitation Learning:} IL trains agents to mimic expert behaviors. Behavior cloning uses supervised learning for replication \cite{kelly2019hg, sasaki2020behavioral, reddysqil, florence2022implicit, shafiullah2022behavior, hoque2023fleet, li2024imitation, mehta2025stable}, while Inverse RL derives reward functions from expert demonstrations \cite{Abbeel2004, Ziebart2008, dadashi2020primal, wang2022adversarially}. Building on IRL, adversarial methods distinguish between expert and learner behaviors to provide reward signals \cite{Ho2016, fu2017learning,li2017infogail, peng2018variational,lee2019efficient, ghasemipour2020divergence}. There are also approaches that aim to integrate the strengths of BC and IRL \cite{watson2024coherent}. Offline IL methods enable robust training without environment interaction \cite{kim2022demodice, xu2022discriminator,ma2022versatile, xu2022a,hong2023beyond,yan2023offline,  li2023mahalo, zhang2023discriminator, sun2023offline}, and strategies addressing dynamic shifts through diverse data have also been proposed \cite{Chae2022}.

\textbf{Cross-Domain Imitation Learning (CDIL):} CDIL transfers expert behaviors across domains with differences in perspectives, dynamics, or morphologies. Approaches include using the Gromov-Wasserstein metric for cross-domain similarity rewards \cite{fickinger2022gromov}, timestep alignment \cite{sermanet2018, Kim2020, Liu2018, Raychaudhuri2021}, and temporal cycle consistency to address alignment issues \cite{Zakka2022}. Techniques also involve removing domain-specific information via mutual information \cite{Cetin2021}, maximizing transition similarity \cite{Franzmeyer2022}, or combining cycle consistency with mutual information \cite{yin2022cross}. Adversarial networks and disentanglement strategies further enhance domain invariance \cite{Stadie2017, Sharma2019, shang2021, choi2024domain}.

\textbf{Imitation from Observation (IfO):} IfO focuses on learning behaviors without access to action information. Approaches can be divided into those leveraging vectorized observations provided by the environment \cite{Torabi2018generative, zhu2020off,desgarat2020, gangwani2022imitation, chang2022flow, liu2023ceil, freund2023coupled} and those utilizing images to model behaviors \cite{li2018oil, liang2018cirl,das2021model, karnan2022voila, karnan2022adversarial, belkhale2023hydra, zhang2024action,xie2024decomposing, ishida2024robust, aoki2024environmental}. Image-based methods, in particular, have gained attention for enabling robots to learn from human behavior captured in images, facilitating tasks like mimicking human actions \cite{sheng2014integrated, yu2018, zhang2022one, mandlekar2023human}.

%==========================================================

\section{Background}
\label{sec:background}

\subsection{Markov Decision Process and RL Setup}

In this paper, all environments are modeled as a Markov Decision Process (MDP) defined by the tuple $\mathcal{M}=(\mathcal{S}, \mathcal{A}, P, R, \gamma, \rho_0)$, where $\mathcal{S}$ is the state space, $\mathcal{A}$ the action space, $P: \mathcal{S} \times \mathcal{A} \times \mathcal{S} \to \mathbb{R}^+$ the state transition probability, $R: \mathcal{S} \times \mathcal{A} \to \mathbb{R}$ the reward function, $\gamma \in (0, 1)$ the discount factor, and $\rho_0$ the initial state distribution. At each timestep $t$, the agent selects an action $a_t \sim \pi$ from a stochastic policy $\pi: \mathcal{S} \times \mathcal{A} \to \mathbb{R}^+$. The environment provides a reward $r_t = R(s_t, a_t)$ and the next state $s_{t+1} \sim P(\cdot|s_t, a_t)$. The goal in reinforcement learning (RL) is to optimize the policy $\pi$ to maximize the discounted cumulative reward $\sum_t \gamma^t r_t$.

\subsection{Adversarial Imitation Learning}
IL trains a learner policy $\pi^L$ to mimic an expert policy $\pi^E$ using an offline dataset $\mathcal{B}^E$ of expert trajectories $\tau^{\pi^E}$, where each trajectory $\tau^\pi := (s_0, a_0, s_1, a_1, \cdots, s_H)$ consists of state-action pairs, with $a_t \sim \pi(\cdot|s_t)$ for $t=0,\cdots,H-1$, and $H$ is the episode length. To improve IL performance, Generative Adversarial IL (GAIL) \cite{Ho2016} applies GAN \cite{Goodfellow2014} principles to IL by using a label discriminator $F$ to distinguish between learner trajectories $\tau^{\pi^L}$ (label 0) and expert trajectories $\tau^{\pi^E}$ (label 1). Rewards are designed in an IRL framework such that $F$ assigns higher rewards to actions that are more ambiguous to classify. Here, the learner $\pi^L$ acts as a generator, aiming to confuse $F$ by performing online RL to maximize these rewards, engaging in adversarial training to align the trajectory distributions of $\pi^L$ and $\pi^E$. Building on this framework, Adversarial IRL (AIRL) \cite{fu2017learning} introduces a reward structure designed to enhance the learner's ability to perform IL more effectively, as follows:
\vspace{-2em}

{\small
\[R_F(s_t,a_t, s_{t+1})=\log F(s_t,a_t,s_{t+1})-\log(1-F(s_t,a_t,s_{t+1}))\]}

\subsection{Cross-Domain IL with Visual Observations}

To enable practical IL in cross-domain scenarios, the expert's environment is modeled as an MDP $\mathcal{M}^S$ in the source domain $S$, and the learner's as an MDP $\mathcal{M}^T$ in the target domain $T$. The goal is to train the learner for cross-domain IL by minimizing domain differences and mimicking expert behavior through distribution matching techniques \cite{Torabi2018generative, gangwani2022imitation, liu2023ceil}. In real-world applications, image observations in offline datasets are often used \cite{Stadie2017, Kim2020, Zakka2022}. The observation space $\mathcal{O}^d$ is part of $\mathcal{M}^d$ for each domain $d \in \{S, T\}$, where each image frame $o_t^d$ captures a snapshot at time $t$. Since single frames cannot capture dynamics, IL relies on sequences of $L$ frames, $o_{\mathrm{seq},t}^d = (o_{t-L+1}^d, \cdots, o_t^d)$, with $L=4$ fixed in this work.

Recent cross-domain IL methods leverage random policies to capture domain characteristics \cite{Cetin2021, choi2024domain}. Here, we define $\pi^{SE}$ as the source expert (SE) policy, $\pi^{SR}$ as the source random (SR) policy, $\pi^{TL}$ as the target learner (TL) policy, and $\pi^{TR}$ as the target random (TR) policy. Offline datasets $\mathcal{B}^\pi$, consisting of visual demonstration trajectories $\tilde{\tau}^\pi = (o_0^d, a_0^d, \cdots, o_H^d)$ generated by $\pi \in \{\pi^{SE}, \pi^{SR}, \pi^{TR}\}$, are provided. For simplicity, we denote $\mathcal{B}^{SE}$, $\mathcal{B}^{SR}$, and $\mathcal{B}^{TR}$ as the datasets corresponding to their respective policies. Using these datasets, which lack access to true states, $\pi^{TL}$ is trained to mimic expert behavior in the target domain by reducing the domain gap.

\vspace{-.6em}
\section{Methodology}
\label{sec:method}

\vspace{-.2em}
\subsection{Domain-Invariant Per-Frame Feature Extraction}
\vspace{-.2em}

In this section, we propose a domain-invariant per-frame feature extraction (DIFF) method to eliminate domain-specific information while preserving expertise-related details, enabling effective domain adaptation prior to utilizing image sequences for expertise assessment. Specifically, we define a shared encoder $p$ and domain-specific decoders $q^d$, where $d \in \{S, T\}$. The encoder $p$ encodes image data into latent features $z^d_t \sim p(\cdot | o^d_t)$, while each decoder $q^d$ reconstructs the original image as $\hat{o}^d_t = q^d(z^d_t)$. This ensures that the feature $z^d_t$ captures essential image characteristics. However, $z^d_t$ may still contain irrelevant domain-specific details (e.g., background, camera angles) in addition to expertise-related information (e.g., agent position, joint angles), which can hinder the learner’s ability to interpret expertise.

To address residual domain-specific details in latent features $z^d_t$, we employ a Wasserstein GAN (WGAN) \cite{gulrajani2017improved}, where the encoder $p$ acts as the generator and a frame discriminator $D_f$ distinguishes whether $z^d_t$ originates from the source or target domain. The frame discriminator $D_f$ is trained to assign higher scores to source features $z^S_t$ and lower scores to target features $z^T_t$, maximizing its ability to classify domains. Conversely, the encoder $p$ learns in the opposite direction, aiming to confuse $D_f$. This adversarial process aligns the distributions of $z^d_t$ across domains, effectively removing domain-specific information in $z^d_t$. Simultaneously, the encoder-decoder structure preserves task-relevant details by penalizing reconstruction errors, ensuring $z^d_t$ retains expertise-critical information while achieving domain invariance. To further enhance alignment, we incorporate a consistency loss inspired by \cite{Zhu2017}, ensuring $z^d_t$ retains expertise-related information even when transferred between domains \cite{choi2024domain}. Specifically, when $z^d_t$ passes through the opposite domain’s decoder $q^{d'}$ and is re-encoded by $p$, the resulting latent $\hat{z}^d_t \sim p(\cdot | q^{d'}(z^d_t))$ remains consistent. This process removes domain-specific artifacts while preserving task-relevant features. The training involves three components: the frame discriminator loss $\mathcal{L}_{\textrm{disc},f}(D_f)$, the frame generator loss $\mathcal{L}_{\textrm{gen},f}(p)$, and the encoder-decoder loss $\mathcal{L}_{\textrm{enc-dec}}(p,q)$, defined as:
\vspace{-2em}

{\small
\begin{align}
&\mathcal{L}_{\textrm{disc},f}:= \mathbb{E}_{\substack{\small z^S_t \sim p(\cdot|o^S_t),\\\small z^T_t \sim p(\cdot|o^T_t)}}\left[-D_f(z^S_t)+D_f(z^T_t)\right] + \lambda_{\mathrm{gp},f} \cdot GP,\nonumber\\
&\mathcal{L}_{\textrm{gen},f}:= \mathbb{E}_{\substack{\small z^S_t \sim p(\cdot|o^S_t),\\\small z^T_t \sim p(\cdot|o^T_t)}}\left[D_f(z^S_t)-D_f(z^T_t)\right],\\
&\mathcal{L}_{\textrm{enc-dec}}:= \sum_{d=S,T}\mathbb{E}_{z_t^d\sim p(\cdot|o^d_t)}\big[\underbrace{\| o^d_t - \hat{o}^d_t \|_2 
}_{\text{Reconstruction Loss}} + \underbrace{\| \bar{z_t} - \hat{z}^d_t \|_2}_{\text{Feature Consistency Loss}}\big]\nonumber,
\end{align}}
where $\hat{o}^d_t = q^d(z^d_t)$, $\hat{z}^d_t \sim p(\cdot | q^{d'}(z^d_t))$, and $\bar{x}$ represents stopping gradient flow for $x$ and $GP$ represents the Gradient Penalty term to guarantee stable learning. Samples are drawn from domain-specific buffers $\mathcal{B}^S := \mathcal{B}^{SR} \cup \mathcal{B}^{SE}$ and $\mathcal{B}^T := \mathcal{B}^{TR} \cup \mathcal{B}^{TL}$, where $\mathcal{B}^{TL}$ stores trajectories from $\pi^{TL}$ during training. This approach ensures domain-invariant features while preserving task-relevant information, enabling robust cross-domain expertise alignment. 

\begin{figure}[!t]
    \begin{center}
    \centerline{\includegraphics[width=1\columnwidth]{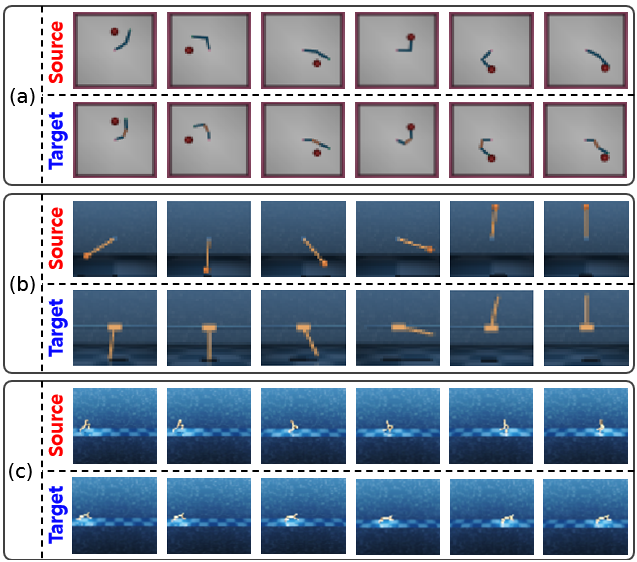}}
    \vspace{-.5em}
    \caption{Image mappings of DIFF based on aligned latent features in (a) Reacher, (b) Pendulum, and (c) MuJoCo tasks.}
    \vspace{-4em}
    \label{fig:frametmap}
    \end{center}
\end{figure}

Fig. \ref{fig:frametmap} showcases image mappings with the proposed DIFF, aligning source and target images based on their closest latent features across (a) Reacher, guiding a robotic arm to target points; (b) Pendulum, swinging up and balancing upright; and (c) MuJoCo, moving forward rapidly despite differing dynamics. Each environment features distinct agents in the source and target domains but shares common goals. As shown, images exhibit near one-to-one alignment, capturing task-specific features (e.g., pole angles, agent position) while minimizing domain-specific differences (e.g., appearance, physics), demonstrating the method's ability to retain expertise-related details for accurate behavior mimicking.

\subsection{Sequential Matching with Expertise Labeling}

The proposed per-frame feature extraction removes domain-specific information from individual frames. Building on this, we utilize sequences of these features, as in existing AIL methods, for expertise assessment. At each time step $t$, the feature sequence is defined as $z_{\mathrm{seq},t}^d := {z_{t-L+1}^d, \cdots, z_t^d}$, with $L$ as the fixed sequence length. To classify expertise, we introduce a sequence label discriminator $F_{\textrm{label},s}(z_{\mathrm{seq},t}^d) \in [0,1]$, trained to label feature sequences $z_{\mathrm{seq},t}^d$ from $\mathcal{B}^{SE}$ as expert (label 1) and others as non-expert (label 0). Although per-frame domain-specific information is removed, domain-specific sequence differences (e.g., speeds, step sizes) may persist. To address this, we extend WGAN to feature sequences using a sequence discriminator $D_s$, ensuring residual sequence-related domain-specific information is further eliminated. In summary, the training for feature sequences also includes three components: sequence discriminator loss $\mathcal{L}_{\textrm{disc},s}(D_s)$, sequence generator loss $\mathcal{L}_{\textrm{gen},s}(p)$, and sequence label loss $\mathcal{L}_{\textrm{label},s}(F_{\textrm{label},s},p)$, are defiend as:
\vspace{-2em}

{\small
\begin{align}
&\mathcal{L}_{\textrm{disc},s}:= \mathbb{E}_{\substack{\small z^S_{\mathrm{seq},t} \sim p(\cdot|o^S_{\mathrm{seq},t}),\\\small z^T_{\mathrm{seq},t} \sim p(\cdot|o^T_{\mathrm{seq},t})}}\left[-D_s(z^S_{\mathrm{seq},t})+D_s(z^T_{\mathrm{seq},t})\right]\nonumber \\
&\quad\quad\quad\quad+ \lambda_{\mathrm{gp},s} \cdot GP,\nonumber\\
&\mathcal{L}_{\textrm{gen},s}:= \mathbb{E}_{\substack{\small z^S_{\mathrm{seq},t} \sim p(\cdot|o^S_{\mathrm{seq},t}),\\\small z^T_{\mathrm{seq},t} \sim p(\cdot|o^T_{\mathrm{seq},t})}}\left[D_s(z^S_{\mathrm{seq},t})-D_s(z^T_{\mathrm{seq},t})\right] \\
&\mathcal{L}_{\textrm{label},s}:=\sum_{d=S,T}\mathbb{E}_{z_{\mathrm{seq},t}^d\sim p}\big[\underbrace{\text{BCE}(F_{\textrm{label},s}(z_{\mathrm{seq},t}^d),\mathbbm{1}_{o_{\mathrm{seq},t}^d\sim \mathcal{B}^{SE}}) 
}_{\text{Label Loss}} \big]\nonumber,
\end{align}}
where BCE is the Binary Cross Entropy and $\mathbbm{1}_x$ is the indicator function, which is $1$ if the condition $x$ is true and $0$ otherwise. The proposed loss ensures that the feature sequence $z_{\mathrm{seq},t}^d$ is free from domain-specific information, enabling pure expertise assessment. For WGAN, to balance per-frame and sequence mappings, the unified WGAN loss is redefined as:
$\mathcal{L}_{\textrm{WGAN}} = \lambda_{\textrm{disc}} \mathcal{L}_{\textrm{disc}}+\lambda_{\textrm{gen}}\mathcal{L}_{\textrm{disc}},$
where $\lambda_{\textrm{disc}}$ and $\lambda_{\textrm{gen}}$ are the scaling coefficients for discriminator and generator losses, respectively. The losses are defined as:
\begin{align}
\mathcal{L}_{\textrm{disc}}&:=\alpha \mathcal{L}_{\textrm{disc},f} + (1-\alpha) \mathcal{L}_{\textrm{disc},s}\nonumber\\
\mathcal{L}_{\textrm{gen}}&:=\alpha \mathcal{L}_{\textrm{gen},f} + (1-\alpha) \mathcal{L}_{\textrm{gen},s}.\nonumber
\end{align}
where $\alpha \in (0,1)$ is the WGAN control parameter, adjusting the balance between per-frame WGAN ($\mathcal{L}_{\textrm{disc},f}$, $\mathcal{L}_{\textrm{gen},f}$) and sequence WGAN ($\mathcal{L}_{\textrm{disc},s}$, $\mathcal{L}_{\textrm{gen},s}$). Due to the large number of losses, most scales are fixed, while parameter search is conducted for the key WGAN hyperparameters $\alpha$, $\lambda_{\textrm{disc}}$, and $\lambda_{\textrm{gen}}$, which are most relevant to the proposed DIFF method. Details on other loss scales are provided in Appendix \ref{secapp:imp}.

\subsection{Frame-wise Time Labeling and Reward Design}

The trained $F_{\textrm{label},s}$ evaluates the expertise of feature sequences, with labels influenced by the overlap between the source domain's expert data $\mathcal{B}^{SE}$ and random data $\mathcal{B}^{SR}$ during sequence label loss training. When expert and random sequences overlap significantly, expert labels are distributed between 0 and 1, helping the target learner distinguish and mimic critical behaviors effectively. However, as shown in Fig. \ref{fig:motiv}(b), minimal overlap results in most expert data being labeled as 1 when slightly deviating from random data, making it harder to identify important behaviors. To address this, we propose a frame-wise time labeling method, which segments expert behavior by timesteps and guides the learner to prioritize frames from later timesteps to achieve task objectives. To implement the frame labeling, we define additional frame label discriminator $F_{\textrm{label},f}(z_t^d) \in [0,1]$, trained using the frame label loss $\mathcal{L}_{\textrm{label},f}(F_{\textrm{label},f})$:
\vspace{-1em}

{\small
\begin{equation}
\mathcal{L}_{\textrm{label},f}:=\mathbb{E}_{z_t^S\sim p(\cdot|o^S_t)}\left[\text{BCE}\left(  F_{\textrm{label},f}(z_t^S),y_t)\right) \right]\nonumber,
\end{equation}}

\begin{figure}[!t]
\vspace{-0.1in}
    \begin{center}
    \centerline{\includegraphics[width=0.87\columnwidth]{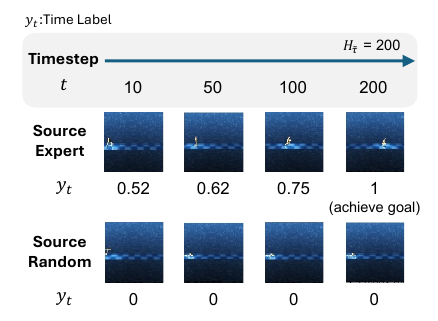}}
    % \centerline{\includegraphics[width=0.82\columnwidth]{motiv_frame.pdf}}
    \vspace{-0.2in}
    \caption{Illustration of frame-wise time labeling}
    \vspace{-0.4in}
    \label{fig:framewise}
    \end{center}
\end{figure}

\vspace{-0.1in}
where $y_t$, the time label for $o_t^S$ at time $t$, is defined as:
\[
y_t = \left\{\begin{array}{cl}\left(\frac{t}{H_{\tilde{\tau}}}+1\right)/2 & \textrm{ if } o_t^S \sim \mathcal{B}^{SE},\\0 &\textrm{otherwise.}\end{array}\right.
\]
Here, $H_{\tilde{\tau}}$ denotes the episode length of $\tilde{\tau} \in \mathcal{B}^{SE}$. Time labeling is trained solely on the source domain, as expert data is unavailable in the target domain. Unlike prior methods aligning features by similar timesteps \cite{sermanet2018}, our approach segments expert behavior with labels for finer granularity in expertise assessment. The frame label discriminator assigns higher labels to later timesteps, segmenting expert behavior and guiding the learner to replicate actions aligned with task objectives. Fig. \ref{fig:framewise} illustrates how time labeling $y_t$ prioritizes later-stage frames, emphasizing the temporal progression of expertise and ensuring precise replication of expert behaviors for accurate task completion.

\begin{figure}[!t]
    \vspace{-0.1in}
    \begin{center}
    \centerline{\includegraphics[width=0.82\columnwidth]{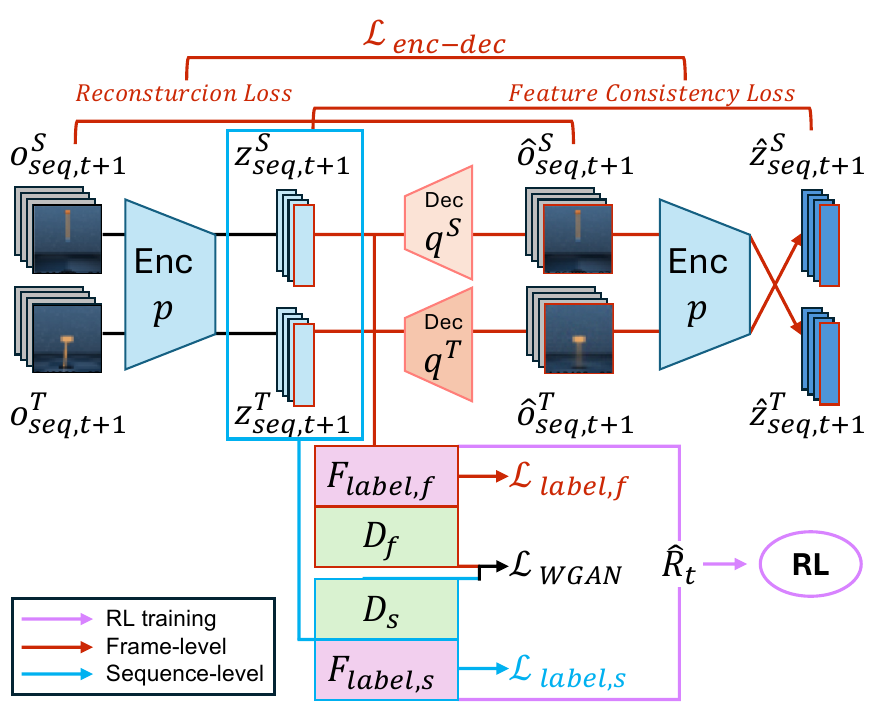}}
    \vspace{-0.15in}
    \caption{Structure of the proposed DIFF-IL}
    \vspace{-3.5em}
    \label{fig:structure}
    \end{center}
\end{figure}

In summary, we propose DIFF for IL (DIFF-IL), which integrates the domain-invariant per-frame feature extraction with AIL principles from Section \ref{sec:background}. DIFF-IL leverages sequence labels via $F_{\textrm{label},s}$ to guide the learner in mimicking expert behavior, while frame-wise time labeling through $F_{\textrm{label},f}$ emphasizes later-stage frames, prioritizing the temporal progression of expertise. DIFF-IL integrates sequence and frame labels into a reward function to maximize their alignment for accurate expert behavior replication:
\[
\hat{R}_t = - \log (1 -  F_{\textrm{label},s} (z_{\mathrm{seq},t+1}^T) \cdot F_{\textrm{label},f}(z_{t+1}^T)),
\]
where $z_{\mathrm{seq},t+1}^T \sim p(\cdot | o_{\mathrm{seq},t+1}^T)$, $z_{t+1}^T \sim p(\cdot | o_{t+1}^T)$ for  $o_{\mathrm{seq},t+1}^T,o_{t+1}^T\sim \mathcal{B}^{TL}$, and observations at time $t+1$ are used in $\hat{R}_t$ to capture the effect of action $a_t$. This reward adopts only the positive part of AIRL's design in Section \ref{sec:background} since it remains effective in maximizing labels. For implementation, the target learner $\pi^{TL}$ aims to maximize the reward sum $\sum_t\gamma^t\hat{R}_t$ using the Soft Actor-Critic (SAC) \cite{Haarnoja2018}, a widely used RL method that exploits entropy for exploration. Each iteration includes $N_{\text{model,train}}$ model training steps and $N_{\text{RL,train}}$ RL training steps, with updates to $p$, $q^S$, $q^T$, $F_{\textrm{label},f}$, and $F_{\textrm{label},s}$ every $n$ periods. More implementation details are provided in Appendix \ref{secapp:imp}, and the DIFF-IL overall structure is summarized in Fig. \ref{fig:structure} and detailed in Algorithm \ref{alg:ours}.

%==========================================================

\vspace{-.6em}
\begin{algorithm}[H]
   \caption{DIFF-IL Framework}
   \label{alg:ours}
\begin{algorithmic}
   \STATE {\bfseries Input:} Source domain data $\mathcal{B}^S$, Target domain data $\mathcal{B}^T$
   \STATE Initialize $p$, $q^S, q^T$, $D_f, D_s$, $F_{\textrm{label},f}, F_{\textrm{label},s}$, $\pi^{TL}$
   
    \FOR{Iteration $i=1$ {\bfseries to} $N_{\text{iter}}$}
       \FOR{Model training step $k=1$ {\bfseries to} $N_{\text{model,train}}$}
           \STATE Sample $(o^S_{\mathrm{seq},t}, o^T_{\mathrm{seq},t}) \sim (\mathcal{B}^S, \mathcal{B}^T)$
           \STATE Calculate $\mathcal{L}_{\textrm{disc},f}$ and $\mathcal{L}_{\textrm{disc},s}$
           \STATE Update $D_f$ and $D_s$ using $\mathcal{L}_{\textrm{disc},f}$ and $\mathcal{L}_{\textrm{disc},s}$
           \IF{$k \mod n = 0$} 
               \STATE Calculate $\mathcal{L}_{\textrm{enc-dec}}$, $\mathcal{L}_{\textrm{gen},f}$, $\mathcal{L}_{\textrm{gen},s}$, $\mathcal{L}_{\textrm{label},f}$, $\mathcal{L}_{\textrm{label},s}$
               \STATE Update $p$, $q^S$, $q^T$, $F_{\textrm{label},f}$, and $F_{\textrm{label},s}$ based on the calculated loss functions
           \ENDIF
       \ENDFOR
       \FOR{RL training step $l=1$ {\bfseries to} $N_{\text{RL,train}}$}
           \STATE Compute reward $\hat{R}_t$ using $F_{\textrm{label},f}$ and $F_{\textrm{label},s}$
           \STATE Perform RL and update the target learner $\pi^{TL}$
       \ENDFOR
       \STATE Store transitions generated by $\pi^{TL}$ in $\mathcal{B}^{TL}$
   \ENDFOR
\end{algorithmic}
\end{algorithm}

\vspace{-2em}

\begin{table*}[!t]
    \centering
    \small
    \resizebox{\textwidth}{!}{%
    \begin{tabularx}{\textwidth}{|c|l|X|X|X|X|X|}
        \hline
        \multirow{2}{*}{} & \multirow{2}{*}{} & \multicolumn{5}{c|}{\textbf{Average Return}} \\ \cline{3-7}
                          &                   & \textbf{DIFF-IL (Ours)} & \textbf{D3IL} & \textbf{DeGAIL} & \textbf{GWIL} & \textbf{TPIL} \\ \hline
        \multirow{8}{*}{\rotatebox{90}{\textbf{Pendulum Tasks}}} 
                          & IP-to-IDP            & \textbf{9358.51 $\pm$ 0.8558} & 9300.23 $\pm$ 271.39 & 479.40 $\pm$ 173.06 & 461.79 $\pm$ 64.80 & 174.52 $\pm$ 52.30 \\
                          & IDP-to-IP            & \textbf{1000.00 $\pm$ 0.00}       & \textbf{1000.00 $\pm$ 0.0}  & 27.00 $\pm$ 192.64 & 417.03 $\pm$ 60.00 & 11.07 $\pm$ 5.84\\
                          & RE3-to-RE2      & -3.33 $\pm$ 0.80            & \textbf{-3.16 $\pm$ 0.73}  & -6.43 $\pm$ 0.70 & -11.57 $\pm$ 0.30 & -9.71 $\pm$1.93 \\
                          & RE2-to-RE3    & \textbf{-2.27 $\pm$ 0.56}   & -3.99 $\pm$ 1.35 & -10.05 $\pm$ 1.24 & -9.84 $\pm$ 0.12 & -10.09 $\pm$ 1.91 \\
                          & Pend-to-CS & \textbf{739.51 $\pm$ 48.54} & 528.54 $\pm$ 106.55 & 1.65 $\pm$ 1.02 & 0.00 $\pm$ 0.21 & 4.70 $\pm$ 14.56 \\
                          & Pend-to-Acrobot  & \textbf{128.24 $\pm$ 40.58}  & 62.51 $\pm$ 28.67 & 3.96 $\pm$ 3.16  & 6.52 $\pm$ 6.46 & 3.77 $\pm$ 4.52 \\
                          & CS-to-Pend & \textbf{803.50 $\pm$ 46.00} & 646.81 $\pm$127.35 & 5.99 $\pm$ 17.48 & 0.21$\pm$ 0.74  & 85.82 $\pm$ 165.01 \\
                          & CS-to-Acrobot  & \textbf{64.86 $\pm$ 25.79}  & 53.97 $\pm$ 30.36 & 2.66 $\pm$ 2.42  & 6.83 $\pm$ 7.81 & 0.54 $\pm$ 1.21  \\\hline
        \multirow{6}{*}{\rotatebox{90}{\textbf{MuJoCo Tasks}}} 
                          & Cheetah-to-Walker    & \textbf{2.84 $\pm$ 0.38}   & 0.12 $\pm$ 0.07 & 0.00 $\pm$ 0.00 & -0.06 $\pm$ 0.06 & 0.00 $\pm$ 0.00 \\
                          & Cheetah-to-Hopper                 & \textbf{1.14 $\pm$ 0.19}            & 0.16 $\pm$ 0.04  & -0.07 $\pm$ 0.07 & -0.02 $\pm$ 0.04 & 0.02 $\pm$0.03 \\
                          & Walker-to-Cheetah      & \textbf{4.54 $\pm$ 0.86}            & 2.78 $\pm$ 0.61  & 0.06 $\pm$ 0.13 & -0.51 $\pm$ 0.13 & 1.38 $\pm$0.24 \\
                          & Walker-to-Hopper            &  \textbf{1.04 $\pm$ 0.28}            & 0.68 $\pm$ 0.11  & 0.00 $\pm$ 0.01 & 0.00 $\pm$ 0.03 & 0.00 $\pm$0.03 \\
                         & Hopper-to-Walker                 & \textbf{2.01 $\pm$ 0.19}            & -0.06 $\pm$ 0.02  & 0.00 $\pm$ 0.01 & -0.03 $\pm$ 0.05 & 0.00 $\pm$0.01 \\
                          & Hopper-to-Cheetah                 & \textbf{2.32 $\pm$ 0.58}            & 0.22 $\pm$ 0.22  & 0.50 $\pm$ 0.37 & 0.33 $\pm$ 0.35 & 0.00 $\pm$0.00  \\\hline
    \end{tabularx}%
    } % End of resizebox
    \vspace{-0.15in}
    \caption{Performance comparison on Pendulum tasks and MuJoCo tasks}
    \vspace{-0.15in}
    \label{tab:perform}
\end{table*}

\vspace{-.3em}
\section{Experiments}
\label{sec:exp}
\vspace{-.2em}

We evaluate the proposed DIFF-IL against various cross-domain IL methods on DeepMind Control Suite (DMC) \cite{tassa2018deepmindcontrolsuite} and MuJoCo \cite{todorov2012}, pairing similar tasks as source and target domains. The evaluation shows how well the target learner mimics the source expert and analyze the effectiveness of key components.

\subsection{Experimental Setup}

\begin{figure}[!t]
    \begin{center}
    \centerline{\includegraphics[width=\columnwidth]{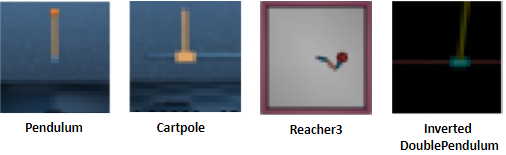}}
    \vspace{-0.1in}
    \caption{Pendulum environments}
    \label{fig:pend_env}
    \vspace{0.05in}
    \centerline{\includegraphics[width=\columnwidth]{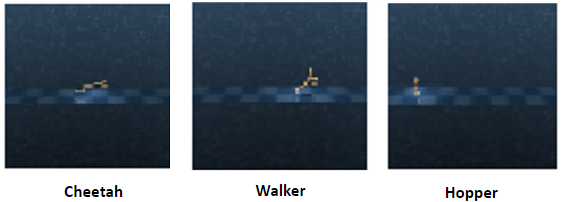}}
    \vspace{-0.15in}
    \caption{MuJoCo environments}
    \vspace{-0.45in}
    \label{fig:mujoco_env}
    \end{center}
\end{figure}

For comparison, we evaluate cross-domain IL methods using images: {\bf TPIL} \cite{Stadie2017}, which extracts domain-invariant features from image sequences; {\bf DeGAIL} \cite{Cetin2021}, which enhances domain information removal with mutual information; {\bf D3IL} \cite{choi2024domain}, which isolates expertise-related behavior using dual consistency loss; and {\bf DIFF-IL} (Ours). Additionally, {\bf GWIL} \cite{fickinger2022gromov}, a state-based approach leveraging Gromov-Wasserstein distance, serves as a baseline. For DIFF-IL, we primarily tuned WGAN hyperparameters ($\alpha$, $\lambda_{\textrm{disc}}$, $\lambda_{\textrm{gen}}$), fixing other loss scales. $\alpha=0.5$ delivered consistently strong performance across environments, while $\lambda_{\textrm{disc}}$ and $\lambda_{\textrm{gen}}$ were optimized per environment. Further experimental details are provided in Appendix \ref{secapp:expdetail}.

\begin{figure*}[t]
        \centering
    \includegraphics[width=0.96\textwidth]{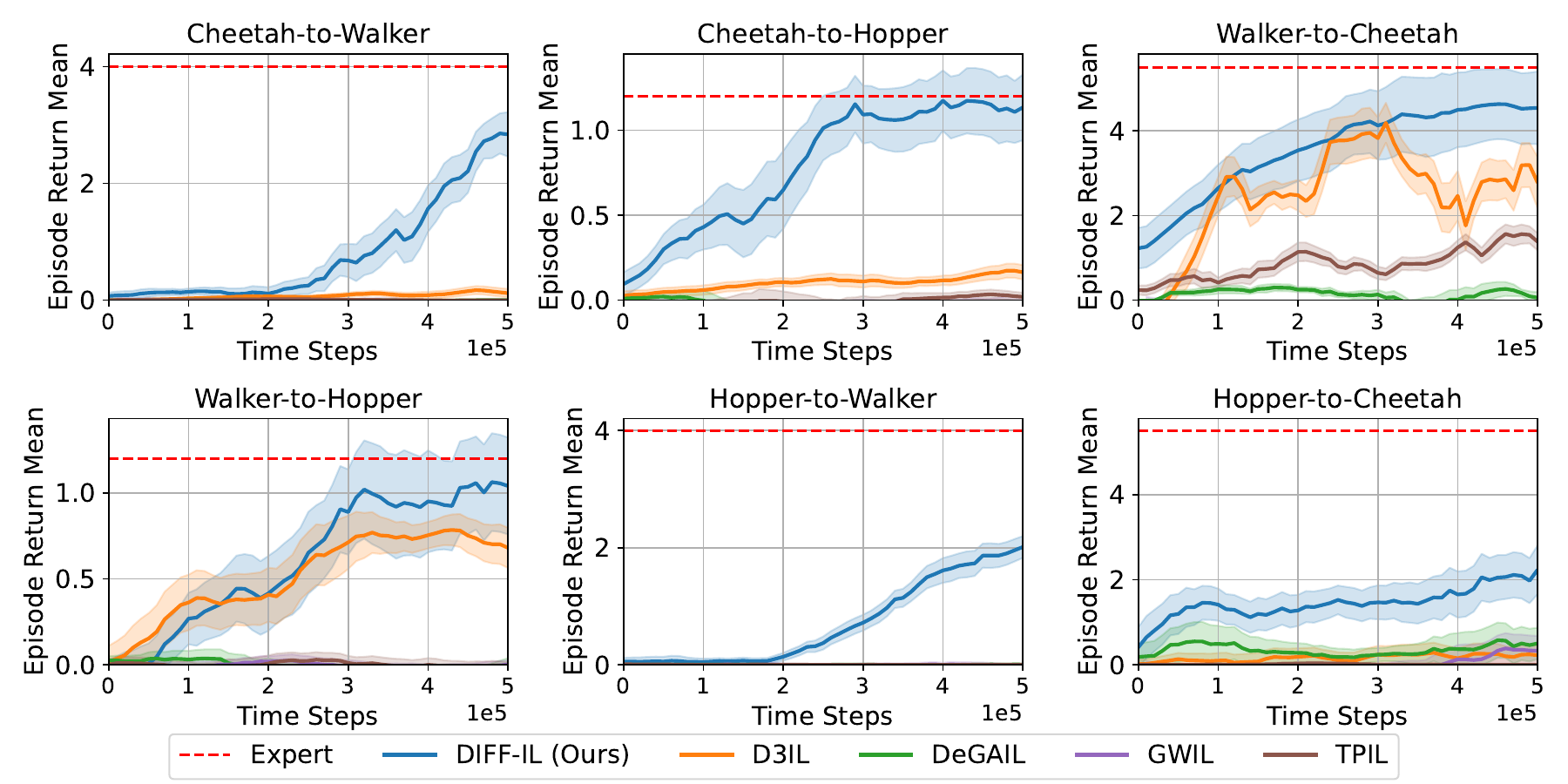}
    \vspace{-0.2in}
    \caption{Performance comparison: Learning curves on MuJoCo}
    \vspace{-0.2in}
    \label{fig:mujgraph}
\end{figure*}

\begin{figure*}[t]
        \centering
    \includegraphics[width=0.94\textwidth]{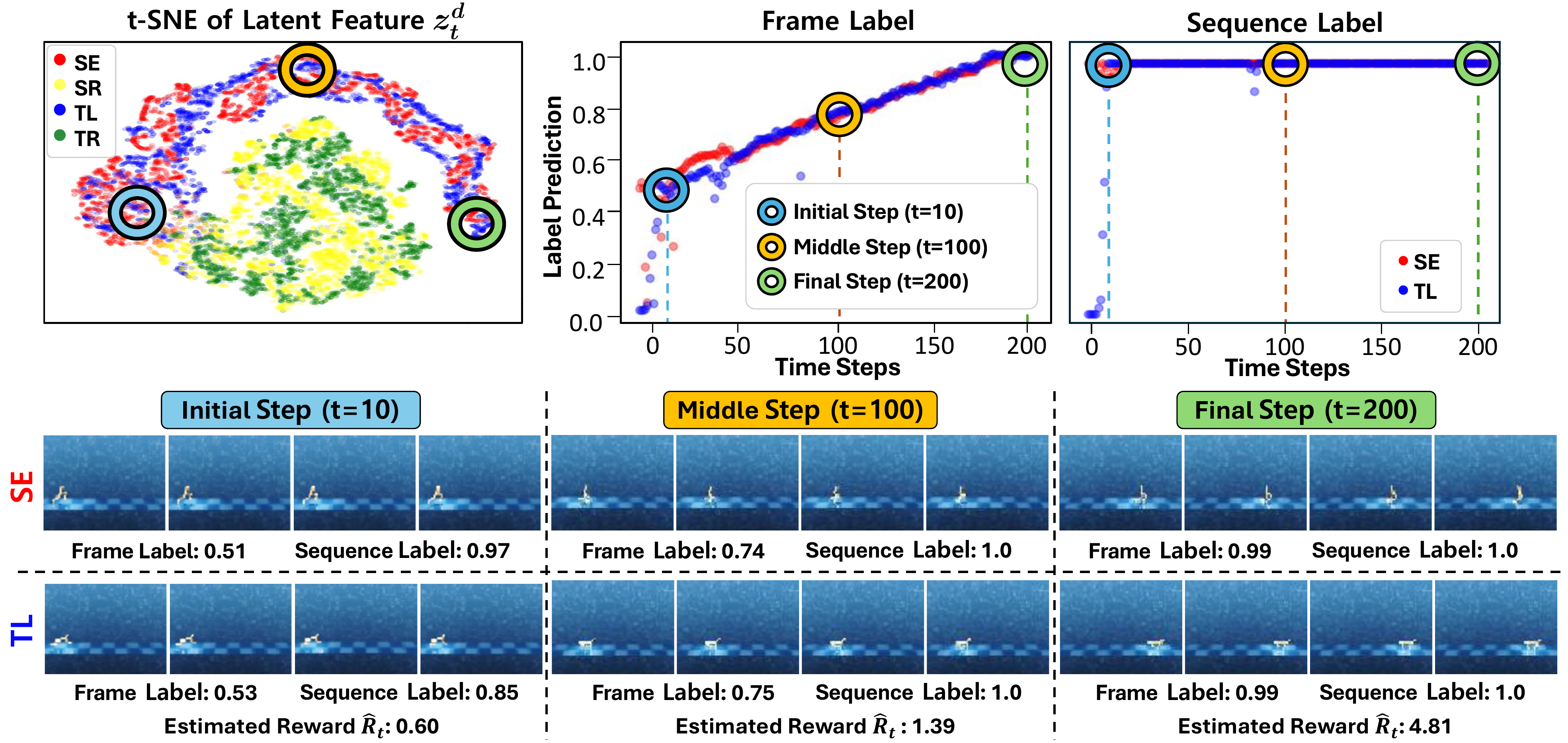}
    \vspace{-0.1in}
    \caption{Image mapping and reward analysis on Walker-to-Cheetah task}
    \vspace{-0.15in}
    \label{fig:visualize}
\end{figure*}

\subsection{Environmental Setup}

We compare the baselines in environments with significant domain differences, focusing on adaptation across tasks with varying agent properties like joints, action spaces, and dynamics, rather than simple changes in viewpoint or color. Cross-domain scenarios pair tasks with similar objectives, denoted as \textit{A-to-B}, where \textit{A} and \textit{B} are the source and target domains. The environments are categorized into \textbf{Pendulum Tasks} and \textbf{MuJoCo Tasks}, and true rewards define task objectives and evaluate the learner's ability to mimic expert behavior but are not used during IL training.

{\bf Pendulum Tasks:} 
Pendulum tasks involve controlling pole agents to maintain balance or reach targets, as shown in Fig.~\ref{fig:pend_env}, and are grouped into three categories: Inverted Pendulum Tasks, including Inverted Pendulum (IP) with a single pole and Inverted Double Pendulum (IDP) with two interconnected poles, focusing on vertical balance with rewards increasing as poles approach an upright position; Reacher Tasks, where Reacher2 (RE2) and Reacher3 (RE3) involve two- and three-joint robotic arms reaching one of 16 targets, with rewards reflecting the negative distance to the target; and DMC Pendulum Tasks, including Cartpole Swingup (CS), Pendulum (Pend), and Acrobot, emphasizing pole balance with rewards increasing for upright positions. Here, DMC Pendulums fix one end of the pole to a central pivot.

{\bf MuJoCo Tasks:} MuJoCo tasks involve locomotion agents, including Walker, Cheetah, and Hopper with distinct state-action spaces, aiming to move as quickly as possible on a given map, as shown in Fig.~\ref{fig:mujoco_env}. The camera is adjusted to clearly capture their movements, with rewards solely based on forward speed, increasing as the agents move faster.

To evaluate domain adaptation, we define 8 pendulum task scenarios (Pend-to-CS, Pend-to-Acrobot, CS-to-Pend, CS-to-Acrobot, RE3-to-RE2, RE2-to-RE3, IP-to-IDP, and IDP-to-IP) and 6 MuJoCo task scenarios (Walker-to-Cheetah, Walker-to-Hopper, Cheetah-to-Walker, Cheetah-to-Hopper, Hopper-to-Cheetah, and Hopper-to-Walker). The Acrobot task, due to its complexity, is used only as a target environment. In each scenario, an expert policy is trained using SAC in the source environment to construct the source expert dataset. Performance is measured as the return achieved by the target learner in the target environment, averaged over 5 random seeds, with results reported as means and standard deviations (shaded areas in graphs and $\pm$ values in tables). Additional details on the environments and offline data construction are provided in Appendix \ref{subsecapp:envsetup}.

\subsection{Performance Comparison}

For various domain adaptation scenarios, Table \ref{tab:perform} presents the mean final return results averaged over the last 10 episodes for all evaluated environments, categorized by method. From the results, it is evident that the proposed algorithm significantly outperforms other cross-domain IL methods in most environments. Notably, Fig. \ref{fig:mujgraph} illustrates the learning curves over time steps in MuJoCo environments. While other algorithms fail to closely replicate expert performance and often struggle to learn effectively, the proposed method successfully mimics expert behavior, enabling the agent to move efficiently and achieve higher scores. However, when Hopper is the source domain, the target performance plateaus below expert levels due to the physical limitations of the Hopper agent, which restrict its maximum achievable speed. As the Hopper expert itself cannot achieve higher speeds, the target domain inherits this limitation, resulting in capped performance. In addition, we also provide detailed learning curves for Pendulum tasks in Appendix \ref{secapp:pend}. These results demonstrate not only superior final performance across most tasks but also significantly faster convergence compared to other cross-domain IL methods, even within the same time steps. This improvement is attributed to the better-designed rewards in the proposed approach. Overall, these comparisons highlight the algorithm’s ability to achieve superior domain adaptation and more effectively mimic the source domain's expert behavior compared to existing methods.

%==========================================================

\subsection{Image Mapping and Reward Analysis}

In IL, understanding how expert behavior is mimicked is as crucial as performance. To analyze how the proposed DIFF-IL effectively mimics a source expert across domains, Fig. \ref{fig:visualize} focuses on the Walker-to-Cheetah environment, where DIFF-IL significantly outperforms other methods. The figure examines the target learner’s (TL) progression toward the goal over time, presenting its image frames at initial, middle, and final timesteps mapped to those of the source expert (SE). Frame and sequence label predictions obtained by the label discriminators $F_{\textrm{label},f}$ and $F_{\textrm{label},s}$ along with the rewards $\hat{R}$, are also visualized. 
For image mapping, each TL frame passes through the encoder $p$, producing a feature $z_t^T$, which is then matched to the SE image in the dataset with the closest feature $z_t^S$, as Fig. \ref{fig:frametmap}. The accompanying t-SNE graph, showing latent features of SE, SR, learned TL, and TR (identical to Fig. \ref{fig:motiv}), illustrates the feature locations and confirms a one-to-one mapping between TL and SE frames across all timesteps. This demonstrates that the learned frame features effectively capture task-relevant positions while excluding domain-specific details.

The t-SNE visualization also reveals minimal overlap between SE and SR data distributions. Consequently, relying solely on sequence labels often misclassifies behaviors slightly deviating from random policies as expert, leading to suboptimal mimicry. In contrast, frame labels finely segment expert behavior over time, assigning higher rewards to frames closer to the goal. This segmentation ensures the agent progresses effectively and aligns its behavior with task objectives. These findings underscore the advantages of the proposed DIFF with frame-wise time labeling, enabling effective goal-oriented learning. Additional analyses, including image mappings and reward evaluations for other environments, are detailed in Appendix \ref{secapp:mapping}, demonstrating similar results consistent with the Walker-to-Cheetah case.

\begin{figure}[!t]
    \begin{center}
    \centerline{\includegraphics[width=\columnwidth]{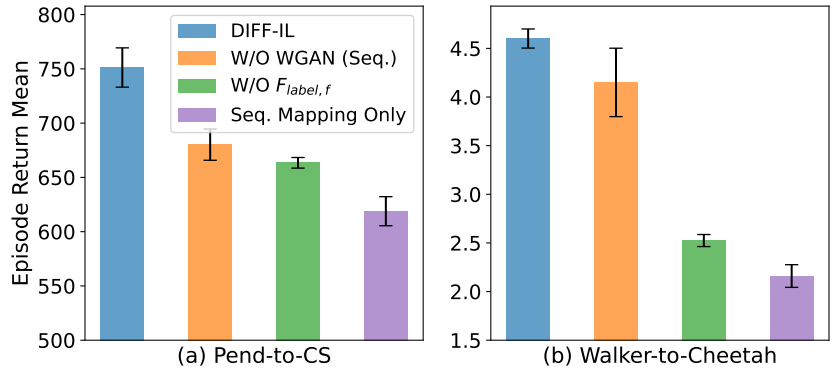}}
    \vspace{-0.1in}
    \caption{Ablation study: Component evaluation}
    \label{fig:comeval}
    \vspace{0.1in}
    \centerline{\includegraphics[width=\columnwidth]{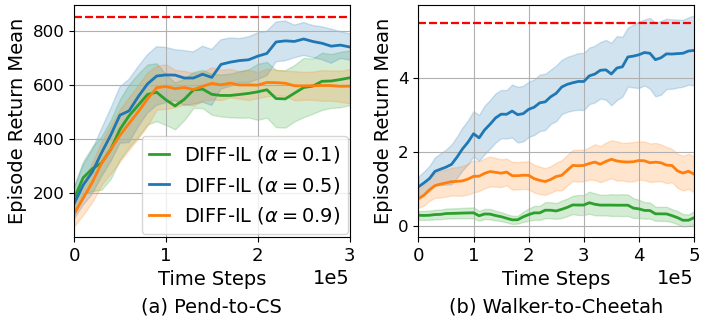}}
    \vspace{-0.1in}
    \caption{Ablation study: WGAN control factor $\alpha$} 
    \vspace{-0.4in}
    \label{fig:abl_alpha}
    \end{center}
\end{figure}
%==========================================================
\subsection{Ablation Studies}

%==========================================================

{\bf Component Evaluation:} To evaluate the components of proposed DIFF-IL, we compare 4 configurations: `W/O WGAN (Seq.)', excluding sequence-based WGAN losses $\mathcal{L}_{\textrm{disc},s}$ and $\mathcal{L}_{\textrm{gen},s}$; `W/O $F_{\textrm{label},f}$', omitting frame-wise time labeling while retaining per-frame feature extraction; `Seq. Mapping Only', using only sequence-based mapping and labeling; and `DIFF-IL', the full method. Fig. \ref{fig:comeval} compares performance in Walker-to-Cheetah and Pend-to-CS, where DIFF-IL shows the superior performance. Result shows that `Seq. Mapping Only' fails to adapt effectively, while `W/O WGAN (Seq.)' and `W/O $F_{\textrm{label},f}$' show moderate improvements. DIFF-IL achieves the highest performance compared to other setups by combining per-frame domain-invariant feature extraction and frame-wise time labeling to prioritize later-stage frames, highlighting their impact on domain adaptation and task success.

{\bf WGAN Control Factor $\alpha$:} To investigate the impact of hyperparameters in DIFF-IL, we conducted an ablation study on WGAN-related hyperparameters. Here, we examine the WGAN control factor $\alpha$, which balances per-frame and sequence-level mapping in DIFF-IL. Fig. \ref{fig:abl_alpha} compares performance in Walker-to-Cheetah and Pend-to-CS for $\alpha=0.1$, $0.5$, and $0.9$. The results indicate that $\alpha=0.5$ achieves the best performance, validating it as the default setting. Lower or higher values reduce performance, highlighting the need for balanced per-frame and sequence-level domain adaptation for effective feature extraction. Additional analyses on $\alpha$ and other hyperparameters across more environments are detailed in Appendix \ref{secapp:moreabl}.

\section{Conclusion}

In this paper, we propose a novel cross-domain IL approach, addressing challenges in image-based observations. Combining per-frame feature extraction with frame-wise time labeling, DIFF-IL successfully removes only domain-specific information. This enables superior alignment and performance, even under significant domain shifts, as demonstrated by experiments showcasing enhanced domain-invariant image mapping and accurate behavior imitation.

% \nocite{langley00}

\section*{Impact Statement}
This paper presents work whose goal is to advance the field of Machine Learning. There are many potential societal consequences of our work, none which we feel must be specifically highlighted here

\bibliography{icml2025}
% \bibliography{example_paper}
\bibliographystyle{icml2025}

%%%%%%%%%%%%%%%%%%%%%%%%%%%%%%%%%%%%%%%%%%%%%%%%%%%%%%%%%%%%%%%%%%%%%%%%%%%%%%%
%%%%%%%%%%%%%%%%%%%%%%%%%%%%%%%%%%%%%%%%%%%%%%%%%%%%%%%%%%%%%%%%%%%%%%%%%%%%%%%
% APPENDIX
%%%%%%%%%%%%%%%%%%%%%%%%%%%%%%%%%%%%%%%%%%%%%%%%%%%%%%%%%%%%%%%%%%%%%%%%%%%%%%%
%%%%%%%%%%%%%%%%%%%%%%%%%%%%%%%%%%%%%%%%%%%%%%%%%%%%%%%%%%%%%%%%%%%%%%%%%%%%%%%
\newpage
\appendix
\onecolumn

\counterwithin{table}{section}
\counterwithin{figure}{section}

\setcounter{equation}{0}
\renewcommand{\theequation}{\thesection.\arabic{equation}}
\section{Detailed Implementation of DIFF-IL}
\label{secapp:imp}

In this section, we detail the implementation of the proposed methods in DIFF-IL. Section \ref{subsecapp:redef} redefines the loss functions, incorporating loss scales and network parameters. Section \ref{subsecapp:impGP} provides the implementation details of the Gradient Penalty (GP) used in WGAN. Section \ref{subsecapp:impRL} explains the implementation of RL losses for training the target learner policy. Section \ref{subsecapp:impNS} details the specific architectures of the networks introduced in Section \ref{subsecapp:redef}, including their structural design and parameter configurations.

\subsection{Redefined Loss Functions for DIFF-IL}
\label{subsecapp:redef}

In this section, we redefine the losses in DIFF-IL, explicitly including their associated parameters, as described in Section \ref{sec:method}. The encoder is parameterized as $\phi$, the domain-specific decoders as $\psi^S$ (source) and $\psi^T$ (target), the frame and sequence discriminators as $\zeta_f$ and $\zeta_s$, and the frame and sequence label discriminators as $\chi_f$ and $\chi_s$, respectively.

The unified WGAN losses for the discriminator and generator are redefined as:
\begin{align}
\label{eq:appdisc}
\mathcal{L}_{\textrm{disc}}(\zeta_f,\zeta_s) &:= \lambda_{\textrm{disc}} \cdot \mathbb{E}\left[\alpha(\underbrace{-D_{\zeta_f}(z^S_t) + D_{\zeta_f}(z^T_t)}_{L_{\textrm{disc},f}(\zeta_f)}) + (1-\alpha)(\underbrace{-D_{\zeta_s}(z^S_{\mathrm{seq},t}) + D_{\zeta_s}(z^T_{\mathrm{seq},t})}_{L_{\textrm{disc},s}(\zeta_s)})\right] + \lambda_{\textrm{gp}} \cdot GP \\ 
\mathcal{L}_{\textrm{gen}}(\phi) &:= \lambda_{\textrm{gen}} \cdot \mathbb{E}\left[\alpha(\underbrace{D_{\zeta_f}(z^S_t) - D_{\zeta_f}(z^T_t)}_{L_{\textrm{gen},f}(\phi)}) + (1-\alpha)(\underbrace{D_{\zeta_s}(z^S_{\mathrm{seq},t}) - D_{\zeta_s}(z^T_{\mathrm{seq},t})}_{L_{\textrm{gen},s}(\phi)})\right] 
\end{align}
where $GP$ is gradient penalty term, $z^d_t \sim p_{\phi}(\cdot|o^d_t)$ and $z^d_{\mathrm{seq},t} \sim p_{\phi}(\cdot|o^d_{\mathrm{seq},t})$ for $d \in \{S, T\}$. The coefficients $\lambda_{\textrm{disc}}$, $\lambda_{\textrm{gen}}$, and $\lambda_{\textrm{gp}}$ control the contributions of the losses, while $\alpha$ balances frame- and sequence-based mappings.

The encoder-decoder loss, incorporating generator, reconstruction, and feature consistency losses, is redefined as:
\begin{equation}
    \mathcal{L}_{\textrm{enc-dec}}(\phi, \psi^S, \psi^T) :=  \sum_{d=S,T} \mathbb{E}_{z_t^d \sim p_{\phi}(\cdot|o_t^d)} \left[\lambda_{\mathrm{recon}}\cdot\underbrace{\| o^d_t - \hat{o}^d_t \|_2}_{\text{Reconstruction Loss}} + \lambda_{\mathrm{fcon}} \cdot\underbrace{\| \bar{z_t} - \hat{z}^d_t \|_2}_{\text{Feature Consistency Loss}}\right], 
\end{equation}
where $d'$ is the opposite domain of $d$, $\hat{o}^d_t = \psi^d(z^d_t)$ and $\hat{z}^d_t \sim p_{\phi}(\cdot | \psi^{d'}(z^d_t))$, with coefficients $\lambda_{\textrm{recon}}$ and $\lambda_{\textrm{fcon}}$ controlling reconstruction and feature consistency losses. The sequence label loss and the frame-wise time labeling loss are redefined as:
\begin{equation}
    \mathcal{L}_{\textrm{label},s}(\phi, \chi_s) := \sum_{d=S,T} \lambda_{\textrm{label},s}^d \cdot\mathbb{E}_{z_{\mathrm{seq},t}^d \sim p_{\phi}(\cdot|o_{\mathrm{seq},t}^d)} \left[\text{BCE}(\mathbbm{1}_{o_{\mathrm{seq},t}^d \sim \mathcal{B}^{SE}}, F_{\chi_s}(z_{\mathrm{seq},t}^d))\right], 
\end{equation}
\begin{equation}
    \mathcal{L}_{\textrm{label},f}(\chi_f) := \lambda_{\textrm{label},f} \cdot \mathbb{E}_{z_t^S \sim p_{\phi}(\cdot|o^S_t)} \left[\text{BCE}(y_t, F_{\chi_f}(z_t^S))\right], 
\end{equation}
where $\lambda_{\textrm{label},s}^d$ is the sequence label loss coefficient and $y_t$ is the time label for frame $o_t^S$. Finally, the reward is redefined as:
\begin{equation}
    \label{equ:reward}
    \hat{R}_t = - \log (1 -  F_{{\chi}_s} (z_{\mathrm{seq},t+1}^T) \cdot F_{\chi_f}(z_{t}^T) + \epsilon)
\end{equation}
where \(\epsilon = 1 \times 10^{-12}\) prevents numerical issues when the product of the sequence and frame labels approaches 1. Details of the loss scale coefficients for all losses are summarized in Appendix \ref{subsecapp:hyper}.

\subsection{Implementation of GP}
\label{subsecapp:impGP}

To ensure stable training of the adversarial network, the WGAN framework \cite{Gulrajani2017} incorporates a gradient penalty (GP) to enforce 1-Lipschitz continuity for the discriminator. In the redefined discriminator loss in Eq. \eqref{eq:appdisc}, the GP term can be defined as follows:

\begin{align}
\text{Gradient Penalty} = \Big(\|\alpha \cdot\nabla_{\delta_{\textrm{label},f}} D_{\zeta_{f}}(\delta_{\textrm{label},f}) 
+ (1-\alpha) \cdot\nabla_{\delta_{\textrm{label},s}} D_{\zeta_{s}}(\delta_{\textrm{label},s})  \|_2 - 1\Big)^2,
\end{align}
where \(\delta_{\textrm{label},f}\) and \(\delta_{\textrm{label},s}\) are the interpolated features between the source and target domain features, computed as:
\begin{align}
\delta_{\textrm{label},f} = \delta z_{t}^{S} + (1 - \delta) z_{t}^{T}, \\
\delta_{\textrm{label},s} = \delta z_{\mathrm{seq}, t}^{S} + (1 - \delta) z_{\mathrm{seq}, t}^{T}.
\end{align}
Here, the frame features \(z_{t}^{S} \sim p_{\phi}(\cdot|o_{t}^{S})\) and \(z_{t}^{T} \sim p_{\phi}(\cdot|o_{t}^{T})\), and the sequence features \(z_{\mathrm{seq}, t}^{S} \sim p_{\phi}(\cdot|o_{\mathrm{seq}, t}^{S})\) and \(z_{\mathrm{seq}, t}^{T} \sim p_{\phi}(\cdot|o_{\mathrm{seq}, t}^{T})\), represent features extracted from the source and target domains, respectively. The scalar \(\delta \sim \text{Unif}(0, 1)\) serves as the interpolation factor. The GP term enforces Lipschitz continuity on the discriminator, stabilizing adversarial training by mitigating extreme gradients and promoting smooth convergence. Additionally, we maintain a 5:1 training ratio between the discriminator and generator, following standard practices to ensure stability  during training.

\subsection{RL Implementation}
\label{subsecapp:impRL}

To train the target learning policy \(\pi^{TL}\), we parameterize both the policy \(\pi^{TL}\) and the state-action value function $Q$ using parameter $\theta$. Utilizing Soft Actor-Critic (SAC) \cite{Haarnoja2018}, the critic and actor losses are defined as follows:
\begin{equation}
\mathcal{L}_Q (\theta) = \mathbb{E}_{(s_t, a_t, s_{t+1}, o_{seq,t}^T) \sim \mathcal{B}^{TL}} \left[ \frac{1}{2}\left( Q_\theta(s_t, a_t) - \left( \hat{R}_t + \gamma \mathbb{E}_{a_{t+1} \sim \pi_\phi(\cdot | s_{t+1})}[Q_{\theta^-}(s_{t+1}, a_{t+1})- \lambda_{\text{ent}} \cdot \text{log}\pi_{\theta}(\cdot|s) ] \right) \right)^2 \right],
\end{equation}

\begin{equation}
\mathcal{L}_\pi (\theta) = \mathbb{E}_{s_t \sim \mathcal{B}^{TL}} \left[ \mathbb{E}_{a_t \sim \pi_\theta(\cdot | s_t)} \left[ D_{KL} \Bigl( \pi_{\theta}(a_t|s_t) || \frac{\text{exp}(Q_{\theta}(s_t,a_t)/\lambda_{\text{ent}})}{Z_{\theta}(s_t)} \Bigr) \right] \right],
\end{equation}
where $D_{KL}$ represents the Kullback-Leibler (KL) divergence, \(Q_\theta(s, a)\) denotes the parameterized state-action value function, $\theta^-$ is the parameter of the target network updated via the exponential moving average (EMA) method, $\pi_\theta(a|s)$ represents the target learner policy parameterized by $\theta$, and $\hat{R}_t$ is computed as in Eq. \ref{equ:reward}, capturing the estimated effect of actions.
The critic loss minimizes the difference between the predicted value $Q_\theta$ and the target value derived from the Soft Bellman equation, ensuring accurate value estimation. The actor loss minimizes the divergence between the policy $\pi_\theta$ and the Softmax distribution induced by $Q$, encouraging the policy to prioritize actions that maximize long-term rewards. To enhance training stability, SAC incorporates double Q-learning and automatic adjustment of the entropy coefficient $\lambda_{\text{ent}}$.

\subsection{Network Architecture and Configurations}
\label{subsecapp:impNS}

This subsection outlines the architecture of the networks used in DIFF-IL, detailing the encoder, decoder, discriminators, label networks, and the SAC-based actor-critic structure, as follows:

\begin{itemize}
\item \textbf{Encoder} ($p_{\phi}$): A convolutional neural network that extracts features from input data. It comprises convolutional layers with 16, 32, and 64 filters, applied with different strides, and utilizes LeakyReLU activations \cite{xu2015empirical} to mitigate vanishing gradient issues. The final output is flattened and passed through a dense layer with 32 units.
\item \textbf{Decoders} ($q^S_{\psi^S}, q^T_{\psi^T}$): Reconstructs data from encoded features using \texttt{ConvTranspose} (transposed convolutional layers) with 64 and 32 filters. It upsamples feature maps back to their original resolution, ending with a \texttt{ConvTranspose} layer outputting a 3-channel image. The final layer uses a linear activation for reconstruction.
\item \textbf{WGAN discriminators} ($D_{\zeta_f}$, $D_{\zeta_s}$): These discriminators distinguish between source and target features from the encoder, operating on either frame or sequence level. They consist of dense layers with LeakyReLU activations and a final dense layer without activation, producing a scalar output indicating whether the input features are from the source or target domain.
\item \textbf{Label discriminators} ($F_{\chi_f}$, $F_{\chi_s}$): Predict labels for frames and sequences using dense layers with LeakyReLU activations. The final layer applies a sigmoid activation to output probabilities for the class labels.
\item \textbf{Critic} ($Q_\theta$): Evaluates the value of actions using dense layers with ReLU activations. The critic outputs the state-action value for each action.
\item \textbf{Target learner policy} ($\pi_\theta$):  Generates actions modeled as independent Gaussian distributions for each action dimension. The policy network outputs  the mean $\mu_\theta$ and standard deviation $\sigma_\theta$, both the mean and standard deviation have sizes equal to the action dimension. This stochastic formulation enables action sampling, facilitating exploration during training.
\end{itemize}
Details about the action dimensions for each environment are available in Appendix \ref{subsecapp:envsetup}, and a summary of the network architecture is presented in Table \ref{app:network_architecture}.

\begin{table*}[h]
\centering
\renewcommand{\arraystretch}{1.25}
\resizebox{\textwidth}{!}{%
\begin{tabular}{|c|c||c|c|}
\hline
\textbf{Network} & \textbf{Layers} & \textbf{Network} & \textbf{Layers} \\
\hline
\multirow{8}{*}{\makecell{\textbf{Encoder} \\ $(p_{\phi})$}} 
& Conv(16, 1, LeakyReLU) & \multirow{7}{*}{\makecell{\textbf{Decoders} \\ $(q^S_{\psi^S},q^T_{\psi^T})$}}
& ConvTranspose(64, 1, LeakyReLU) \\
& Conv(16, 2, LeakyReLU) & & ConvTranspose(64, 2, LeakyReLU) \\
& Conv(32, 1, LeakyReLU) & & ConvTranspose(32, 1, LeakyReLU) \\
& Conv(32, 2, LeakyReLU) & & ConvTranspose(32, 2, LeakyReLU) \\
& Conv(64, 1, LeakyReLU) & & ConvTranspose(16, 1, LeakyReLU) \\
& Conv(64, 2, LeakyReLU) & & ConvTranspose(16, 2, LeakyReLU) \\
& Flatten & & ConvTranspose(3, 1) \\
& Dense(32) & & \\
\hline
\multirow{4}{*}{\makecell{\textbf{WGAN discriminators} \\ $(D_{\zeta_f},D_{\zeta_s})$}} 
& BatchNorm() & \multirow{5}{*}{\makecell{\textbf{Label discriminators} \\ $(F_{\chi_f},F_{\chi_s})$}}    
& BatchNorm() \\
& Dense(400, LeakyReLU) & & Dense(400, LeakyReLU) \\
& Dense(300, LeakyReLU) & & Dense(300, LeakyReLU) \\
& Dense(1) & & Flatten \\
& & & Dense(1, Sigmoid) \\
\hline
\multirow{3}{*}{\makecell{\textbf{Critic} \\ $(Q_\theta)$}} 
& Dense(256, ReLU) & \multirow{3}{*}{\makecell{\textbf{Target learner policy} \\ $(\pi_\theta)$}}           
& Dense(256, ReLU) \\
& Dense(256, ReLU) & & Dense(256, ReLU) \\
& Dense(1) & & Dense(2$\times$Action Dim.) \\
\hline
\end{tabular}%
}
\vspace{-0.1in}
\caption{Architectural specifications of the proposed networks. Conv(nc, stride, act) represents a convolutional layer with nc filters, stride, and activation act. ConvTranspose(nc, stride, act) denotes a transposed convolutional layer. Flatten reshapes the input into a 1D vector. Dense(nc) indicates a dense layer with nc filters.}
\label{app:network_architecture}
\end{table*}

\clearpage
\newpage

\section{Detailed Experimental Setup}
\label{secapp:expdetail}

This section provides the necessary details for conducting the experiments. Section \ref{subsecapp:expsetup} outlines the experimental setup and the design of the ablation study to analyze the impact of key hyperparameters. Section \ref{subsecapp:envsetup} details the environments used in the experiments. Section \ref{subsecapp:hyper} explains the hyperparameters of DIFF-IL and summarizes the optimal configurations. Finally, Section \ref{subsecapp:others} provides a brief overview of the baseline IL algorithms used for performance comparison.

\subsection{Experimental Setup}
\label{subsecapp:expsetup}

Prior to training, we construct datasets essential for the learning process. Buffer sizes for $\mathcal{B}^{SE}, \mathcal{B}^{SR}, \mathcal{B}^{TL},$ and $\mathcal{B}^{TR}$ are fixed at 50K. Among these, $\mathcal{B}^{SE}, \mathcal{B}^{SR},$ and $\mathcal{B}^{TR}$ remain static during training, while $\mathcal{B}^{TL}$ is dynamically updated. After each model and RL training epoch, $\mathcal{B}^{TL}$ is refreshed with 1,000 new samples from environment interactions, replacing the oldest data. Initially, $\mathcal{B}^{TL}$ is populated with random samples, similar to $\mathcal{B}^{TR}$. To construct $\mathcal{B}^{SE}$, we train the source expert policy $\pi^{SE}$ using SAC \cite{Haarnoja2018} and collect samples from $\pi^{SE}$. For $\mathcal{B}^{SR}$ and $\mathcal{B}^{TR}$, random policies are used for data collection. In tasks like IP, IDP, Pendulum, CS, and Acrobot, where random policies can sustain extended downward pole positions, episode lengths vary between expert and random policies. Detailed specifications are provided in Section \ref{subsecapp:envsetup}.

The implementation is based on TensorFlow 2.5 with CUDA 11.4 and CUDNN 8.2.4, running on an AMD EPYC 7313 CPU with an NVIDIA GeForce RTX 3090 GPU. GPU memory usage is approximately 9GB for Pendulum tasks and 18GB for Mujoco tasks, influenced by batch size and feature dimensions. Each epoch requires about one minute. The codebase builds on DeGAIL \cite{Cetin2021}: \url{https://github.com/Aladoro/domain-robust-visual-il}.

\subsection{Environmental Setup}
\label{subsecapp:envsetup}

\begin{figure}[H]
        \centering
    \includegraphics[width=1.0\textwidth]{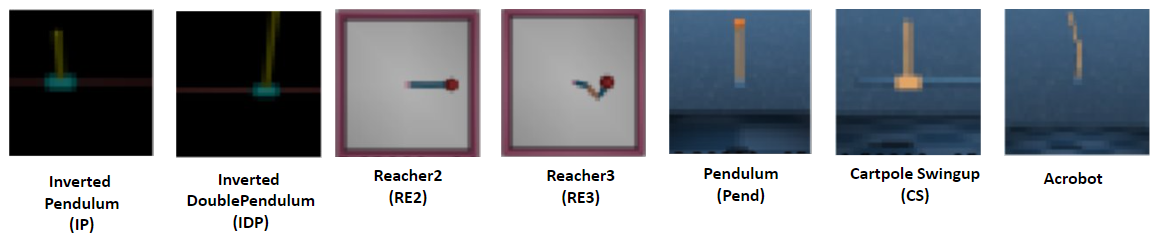}
    \vspace{-0.2in}
    \caption{Image observation of Pendulum environments}
    \vspace{-0.1in}
    \label{app:pendimage}
\end{figure}

\begin{figure}[H]
        \centering
    \includegraphics[width=0.6\textwidth]{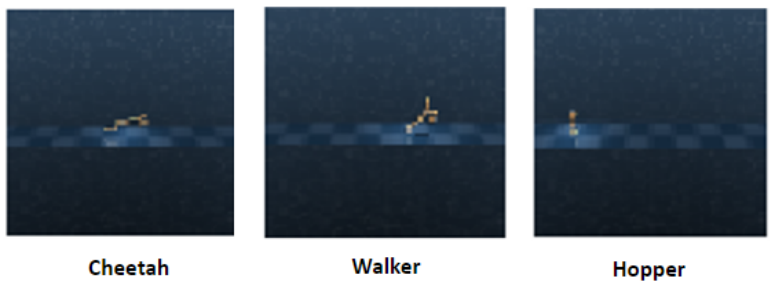}
    \vspace{-0.1in}
    \caption{Image observation of MuJoCo environments}
    \vspace{-0.1in}
    \label{app:mujimage}
\end{figure}

This section outlines the experimental environments categorized into Pendulum Tasks and MuJoCo Tasks. Pendulum Tasks include the Inverted Pendulum (IP) and Inverted Double Pendulum (IDP) from the MuJoCo 150 library, along with modified MuJoCo Reacher environments. Reacher2 (RE2) modifies the goal point of the Reacher environment, while Reacher3 (RE3) extends the arm joint configuration to three joints. Additional environments from the DeepMind Control Suite (DMC) include Pendulum, Cartpole Swingup (CS), and Acrobot. MuJoCo Tasks are adapted DMC environments with a fixed distant camera viewpoint for image-based observations and redesigned reward functions focusing solely on agent velocity. Fig. \ref{app:pendimage} shows image observations for Pendulum environments, and Fig. \ref{app:mujimage} provides those for MuJoCo environments.

\subsubsection{\textbf{Pendulum Tasks}}
\textbf{Inverted Pendulum (IP)}: The Inverted Pendulum task requires balancing a single pole in an upright position. The agent controls the pole’s angle, angular velocity, and the cart's position and velocity. The state space $\mathcal{S}$ is 4-dimensional, while the action space $\mathcal{A}$ is 1-dimensional, representing the force applied to the cart. Rewards increase as the pole remains closer to vertical. Observations are $32 \times 32$ RGB images. Random episodes are $H_{\tilde{\tau}} = 50$, and expert/learner episodes are $H_{\tilde{\tau}} = 1000$.

\textbf{Inverted Double Pendulum (IDP)}: The IDP extends the IP task to two interconnected poles. The state space $\mathcal{S}$ is 11-dimensional, including angles and angular velocities of both poles and the cart’s position and velocity. The action space $\mathcal{A}$ remains 1-dimensional. Rewards increase when both poles are upright. Observations are $32 \times 32$ RGB images. Random episodes are $H_{\tilde{\tau}} = 50$, and expert/learner episodes are $H_{\tilde{\tau}} = 1000$.

\textbf{Reacher Tasks (RE2, RE3)}: These tasks involve controlling a robotic arm with two (RE2) or three (RE3) joints to reach one of 16 randomly assigned targets. The target position is defined in polar coordinates, with $r \in {0.15, 0.2}$ and $\varphi \in {0, \pi/4, \pi/2, \ldots, 7\pi/4}$. The state space $\mathcal{S}$ is 11-dimensional for RE2 and 14-dimensional for RE3, while the action spaces $\mathcal{A}$ have 2 and 3 dimensions, respectively. Negative rewards reflect the distance between the end effector and the target, with zero awarded for reaching the target. Observations are $48 \times 48$ RGB images. For all scenarios, $H_{\tilde{\tau}} = 50$ for random, expert, and learner episodes.

\textbf{Pendulum (Pend)}: The Pendulum task involves balancing a single pole attached to a fixed pivot point. The state space $\mathcal{S}$ is 3-dimensional, and the action space $\mathcal{A}$ is 1-dimensional, representing the torque applied to the pivot. Rewards increase as the pole remains upright. Observations are $32 \times 32$ RGB images. Random episodes are $H_{\tilde{\tau}} = 200$, and expert/learner episodes are $H_{\tilde{\tau}} = 1000$.

\textbf{Cartpole Swingup (CS)}: The CS task requires balancing a pole on a cart moving along a horizontal axis. The state space $\mathcal{S}$ is 5-dimensional, and the action space $\mathcal{A}$ is 1-dimensional, representing the force applied to the cart. Rewards increase when the pole stays upright. Observations are $32 \times 32$ RGB images. Random episodes are $H_{\tilde{\tau}} = 200$, and expert/learner episodes are $H_{\tilde{\tau}} = 1000$.

\textbf{Acrobot}: The Acrobot task involves controlling two connected poles to achieve an upright position from a random position start. The state space $\mathcal{S}$ is 6-dimensional, and the action space $\mathcal{A}$ is 1-dimensional, representing the torque applied to the joint connecting the poles. Rewards increase when the poles reach vertical alignment. Observations are $32 \times 32$ RGB images. Random episodes are $H_{\tilde{\tau}} = 200$, and expert/learner episodes are $H_{\tilde{\tau}} = 1000$.

\subsubsection{\textbf{MuJoCo Tasks}}
\textbf{Cheetah}: The Cheetah environment features a quadrupedal agent designed for fast and efficient running. The state space $\mathcal{S}$ is 17-dimensional, encoding joint angles, velocities, and torso orientation, while the action space $\mathcal{A}$ is 6-dimensional, representing torques applied to joints. Observations are $64 \times 64$ RGB images captured from a fixed camera. The reward function depends solely on forward velocity, aligning with the task's objective. Random and expert/learner episodes are $H_{\tilde{\tau}} = 200$.

\textbf{Walker}: The Walker environment involves a bipedal agent simulating human-like locomotion. Its state space $\mathcal{S}$ is 24-dimensional, including joint angles, velocities, and torso orientation. The action space $\mathcal{A}$ has 6 dimensions, controlling joint torques. Observations are $64 \times 64$ RGB images from a fixed camera. The reward function is modified to depend only on forward velocity. Random and expert/learner episodes are $H_{\tilde{\tau}} = 200$.

\textbf{Hopper}: The Hopper environment tasks a single-legged agent with moving forward efficiently. The state space $\mathcal{S}$ is 15-dimensional, capturing joint positions, velocities, and torso orientation. The action space $\mathcal{A}$ is 4-dimensional, representing joint torques. Observations are $64 \times 64$ RGB images taken from a fixed camera. The reward function relies exclusively on forward velocity, emphasizing efficient locomotion. Random and expert/learner episodes are $H_{\tilde{\tau}} = 200$.

The state dimensions, action dimensions, image sizes, and episode lengths for all environments are summarized in Table \ref{app:tab_envdim}. Image resolution for each task was configured to the minimum level required for clear agent distinction, optimizing memory usage while maintaining sufficient visual detail.

\begin{table*}[h]
  \centering
  \begin{tabular}{c|ccccc}
    \hline
    Environment & 
    $\mathcal{S}$ dim.  & 
    $\mathcal{A}$ dim. &
    Image size &
    Expert Epi. length $(H_{\tilde{\tau}})$&
    Random Epi. length\\
    \hline
    IP          & 4     & 1     & 32x32& 1000& 50 \\
    IDP  & 11    & 1     & 32x32 & 1000& 50\\
    RE2                & 11    & 2     & 48x48& 50& 50\\
    RE3             & 14    & 3     & 48x48& 50& 50\\
    Pend                        & 3     & 1     & 32x32& 1000& 200\\
    CS                        & 5     & 1     & 32x32& 1000& 200\\
    Acrobot                         & 6     & 1     & 32x32& 1000& 200\\
    Cheetah                         & 17     & 6     & 64x64& 200& 200\\
    Walker                         & 24     & 6     & 64x64& 200& 200\\
    Hopper                         & 15     & 4     & 64x64& 200& 200\\
    \hline
  \end{tabular}
  \caption{State, action, and image sizes used in the experiments section.}
  \label{app:tab_envdim}
\end{table*}

\subsection{Hyperparameter Settings}
\label{subsecapp:hyper}

In this subsection, we address the hyperparameters used in the implementation. These hyperparameters are summarized in Tables \ref{tab:shared_parameters}, \ref{app:tab_pendhyper} and \ref{app:tab_mujhyper}. Environment-specific state, action, and image dimensions are in Table \ref{app:tab_envdim}. Task names are abbreviated in the tables for clarity: in Pendulum tasks, 'Pend' is 'P,' and 'Acrobot' is 'A'; in MuJoCo tasks, 'Cheetah' is 'C,' 'Walker' is 'W,' and 'Hopper' is 'H,' with '-' omitted.
As described in the main text, we conducted hyperparameter sweeps only for WGAN-related parameters. All other hyperparameters were fixed at appropriate values, as provided in the tables.

For WGAN-related losses, the discriminator loss weight was searched in the range of 0.01 to 50, while the generator loss weight was searched from 0.01 to 10, with the best hyperparameters selected for each task. Although the search range may seem broad, adjustments were made based on feature mapping quality: when features from the domains did not overlap sufficiently, the discriminator loss was reduced or the generator loss was increased; conversely, when excessive mapping caused the target learner's features to overly align with the expert's, the discriminator loss was increased or the generator loss was reduced. These adjustments are discussed in greater detail in the ablation study in Appendix \ref{secapp:moreabl}.

Also, the WGAN control coefficient $\alpha$, which balances the ratio of frame and sequence mapping, consistently performed best at 0.5 across all tasks and was fixed at this value. For the sequence label discriminator loss scale, the source side was set to a high value of 10, as expert and random data are well separated, while the target side, where label distinctions are less clear, was set to a much smaller value of $1\mathrm{e}{-3}$. 

\begin{table}[h!]
\centering
\begin{tabular}{|c|cc|}
    \hline
    \diagbox{\textbf{shard hyperparameter}}{\textbf{Task}} &\textbf{\large Pendulum tasks} & \textbf{\large MuJoCo tasks} \\ \hline
    \textbf{Reconstruction $(\lambda_{\mathrm{recon}})$} & 0.5 & 1 (HtoW 0.5) \\ \hline
    \textbf{Feature Consistency $(\lambda_{\mathrm{fcon}})$ } & 1 (IPtoIDP, IDPtoIP 0.1) & 1  \\ \hline
    \textbf{Gradient Penalty} $(\lambda_{\mathrm{GP}})$ & 10 & 10 \\ \hline
    \textbf{WGAN control coefficient} $(\alpha)$ & 0.5 & 0.5 \\ \hline
    \textbf{Sequence label Discriminator (Source, $\lambda_{\textrm{label},s}^{S}$)} & 10 & 10 \\ \hline
    \textbf{Sequence label Discriminator (Target, $\lambda_{\textrm{label},s}^{T}$)} & 1e-3 & 1e-3\\ \hline
    \textbf{Frame Labelnet} $\lambda_{\textrm{label},f}$ & 10& 10 \\ \hline
    \textbf{Optimizer} & 1e-3 & 1e-3 \\ \hline
\end{tabular}
\caption{Shared hyperparameters across all tasks. (tasks, value) next to each value indicates exception tasks.}
\label{tab:shared_parameters}
\end{table}

\begin{table}[h!]
    \centering
    \resizebox{\textwidth}{!}{%
    \begin{tabular}{|c|cccccccc|}
        \hline
        \diagbox{\textbf{Hyperparameter}}{\textbf{Task}} & \textbf{\small IPtoIDP} & \textbf{\small IDPtoIP} & \textbf{\small RE2toRE3} & \textbf{\small RE3toRE2} & \textbf{\small PtoCS} & \textbf{\small PtoA} & \textbf{\small CStoP} & \textbf{\small CStoA} \\ \hline
        \textbf{Discriminator $(\lambda_{\mathrm{disc}})$} & 1 & 1 & 1 & 50 & 50 & 1 & 50 & 50 \\ \hline
        \textbf{Generator $(\lambda_{\mathrm{gen}})$} & 0.05 & 0.05 & 1 & 1 & 0.5 & 10 & 0.5 & 10 \\ \hline
        \textbf{Model batch size} & 128 & 128 & 64 & 64 & 128 & 128 & 128 & 128 \\ \hline
        \textbf{Model train num} & 200 & 200 & 100 & 100 & 100 & 100 & 100 & 100\\ \hline
        \textbf{RL train num} & 2000 & 2000 & 2000 & 2000 & 1000 & 1000 & 1000 & 1000 \\ \hline
    \end{tabular}
    } % End of resizebox
    \caption{Hyperparameter setup for Pendulum tasks }
    \label{app:tab_pendhyper}
\end{table}

\clearpage
\newpage

\begin{table}[h!]
    \centering
    % \resizebox{0.9\textwidth}{!}{%
    \begin{tabular}{|c|cccccc|}
        \hline
        \diagbox{\textbf{Hyperparameter}}{\textbf{Task}} & \textbf{WtoC} & \textbf{CtoW} & \textbf{HtoC} & \textbf{CtoH} & \textbf{WtoH} & \textbf{HtoW} \\ \hline
        \textbf{Discriminator $(\lambda_{\mathrm{disc}})$} & 0.5 & 0.02 & 0.1 & 0.05 & 1 & 0.02 \\ \hline
        \textbf{Generator $(\lambda_{\mathrm{gen}})$} & 1 & 0.05 & 0.1 & 0.01 & 0.01 & 0.05 \\ \hline
        \textbf{Model batch size} & 64 & 64 & 64 & 64 & 64 & 64 \\ \hline
        \textbf{Model train num} & 100 & 100 & 100 & 100 & 50 & 50 \\ \hline
        \textbf{RL train num} & 1000 & 1000 & 1000 & 1000 & 1000 & 1000 \\ \hline
    \end{tabular}
    % } % End of resizebox
    \caption{Hyperparameter setup for MuJoCo tasks }
    \label{app:tab_mujhyper}
\end{table}

\subsection{Other Cross-Domain IL Methods}
\label{subsecapp:others}

In this section, we briefly describe the approaches of the four cross-domain algorithms compared with our method:

\textbf{TPIL} \citep{Stadie2017} addresses domain shift in imitation learning by combining unsupervised domain adaptation \citep{Ganin2015} with GAIL \citep{Ho2016}. It uses an encoder to extract domain-independent features, a domain discriminator to differentiate domains, and a label discriminator to classify expert and non-expert behaviors. A gradient reversal layer optimizes these components simultaneously, aligning features across domains for effective policy learning. Code: \url{https://github.com/bstadie/third_person_im}.

\textbf{DeGAIL} \cite{Cetin2021} extracts domain-free features by reducing mutual information between source and target domain data passed through the same encoder. It trains the encoder to minimize domain-related information while using GAIL for reward estimation and reinforcement learning. Code: \url{https://github.com/Aladoro/domain-robust-visual-il}.

\textbf{GWIL} \cite{fickinger2022gromov} leverages the Gromov-Wasserstein distance \cite{memoli2011gromov} as a direct reward, learning optimal coupling between expert and imitator state-action spaces. This distance measures action similarity and guides policy optimization via policy gradient methods. Code: \url{https://github.com/facebookresearch/gwil}.

\textbf{D3IL} \cite{choi2024domain} enhances feature extraction using dual encoders for domain-specific and behavior-specific features, with discriminators refining extraction accuracy through cycle-consistency and reconstruction. A discriminator generates rewards by distinguishing between expert and learner behaviors. Code: \url{https://github.com/sunghochoi122/D3IL}.

For the RL implementation of the target learner policy, all cross-domain IL algorithms are implemented using SAC \cite{Haarnoja2018}. Although TPIL originally employs TRPO \cite{schulman2017trpo}, we re-implemented it using SAC to ensure a fair comparison, following the approach suggested in the D3IL paper \cite{choi2024domain}.

\clearpage
\newpage

\section{Learning Curves on Pendulum Tasks}
\label{secapp:pend}

\begin{figure}[!h]
        \centering
    \includegraphics[width=1.0\textwidth]{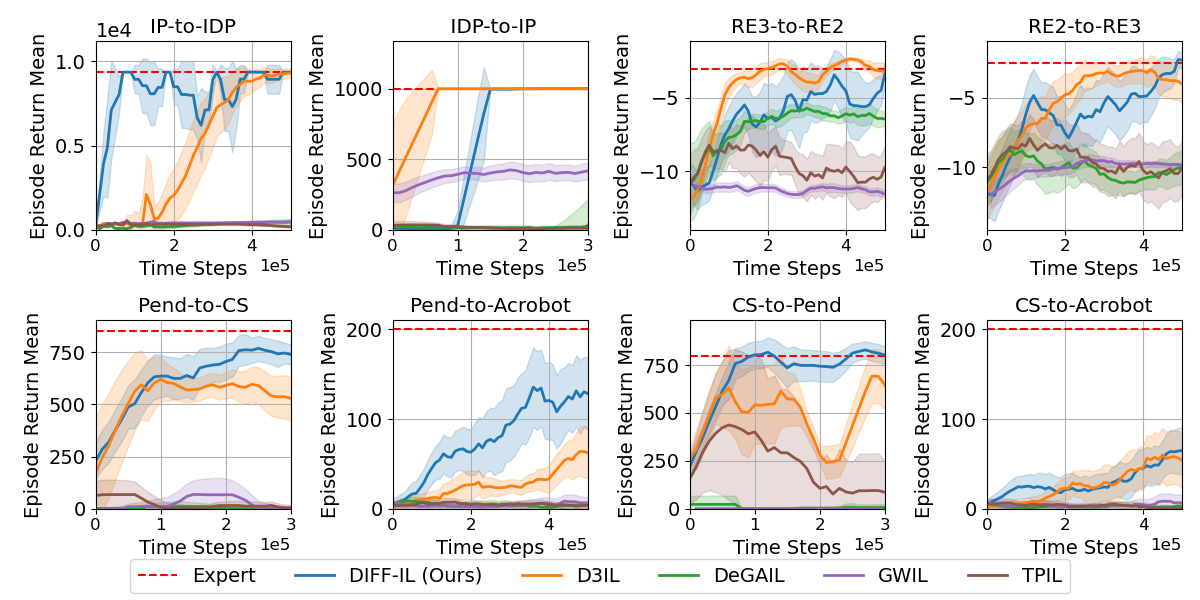}
    \vspace{-0.1in}
    \caption{Performance comparison: Learning curves on Pendulum tasks}
    \vspace{-0.1in}
    \label{app:pendgraph}
\end{figure}

This section presents the learning curves for Pendulum Tasks not covered in Sec. \ref{sec:exp}, with timesteps allocated based on the learning difficulty of each target environment. 
As shown in Fig. \ref{app:pendgraph}, the proposed DIFF-IL demonstrated strong performance across most of the compared environments. Notably, the proposed method excels in DMC Pendulum tasks, showing a significant advantage in environments such as Pend-to-CS, Pend-to-Acrobot, and CS-to-Pend. It achieves faster convergence and significantly outperforms other cross-domain IL methods. This demonstrates that the proposed algorithm designs rewards more effectively for mimicking expert behavior compared to other IL approaches, aligning with the results presented in the main text.

\newpage
\section{Image Mapping and Reward Analysis for Remaining Tasks}
\label{secapp:mapping}

This section explains the feature alignment for tasks not covered in Sec. \ref{sec:method}. For each task-specific figure, images are aligned by processing the Target Learner (TL) and Source Expert (SE) data through the encoder to extract domain-invariant features. The closest features between SE and TL are then matched for alignment. The images are arranged sequentially from left to right, showing the progression of timesteps. The bottom row displays the changes in Target Random (TR) data over time.

Below each image, the estimated frame and sequence label values generated by the frame label discriminator $F_{\textrm{label},f}$ and sequence label discriminator $F_{\textrm{label},s}$ are provided. For TL data, the estimated rewards $\hat{R}_t$ calculated using the proposed reward estimation method are also shown. These visualizations highlight how the model achieves alignment and how the discriminators contribute to effective feature mapping and reward assignment throughout the task.

\subsection{Pendulum Tasks}

\begin{figure}[!h]
    \begin{center}
    \centerline{\includegraphics[width=0.9\columnwidth]{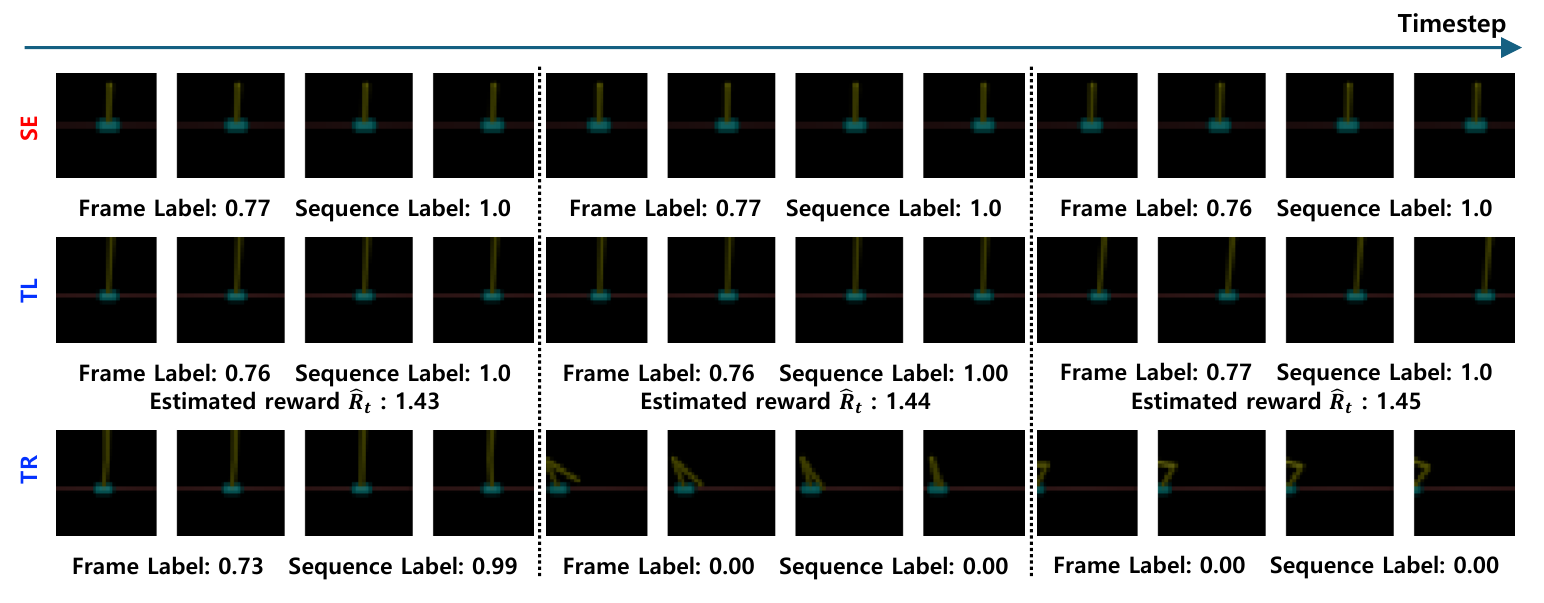}}
    \caption{Image mapping and reward analysis on IP-to-IDP task}
    \label{app:map_iptoidp}
    \end{center}
\end{figure}

\textbf{IP-to-IDP task}: Figure \ref{app:map_iptoidp} illustrates the IP-to-IDP task, highlighting image mapping, label estimation, and reward analysis. In this task, the pole starts upright, and the goal is to maintain balance throughout the episode. 
The TL data effectively aligns with SE frames across timesteps, demonstrating successful learning and accurate domain-invariant feature extraction. Initially, TR samples receive similar rewards to TL due to comparable states, but as the pole begins to fall, frame ($F_{\textrm{label},f}$) and sequence ($F_{\textrm{label},s}$) label values for TR rapidly decline to near zero. This analysis validates the proposed method’s ability to reward expert-like behaviors and penalize deviations, effectively enabling robust learning in complex tasks.

\begin{figure}[!h]
    \begin{center}
    \centerline{\includegraphics[width=0.9\columnwidth]{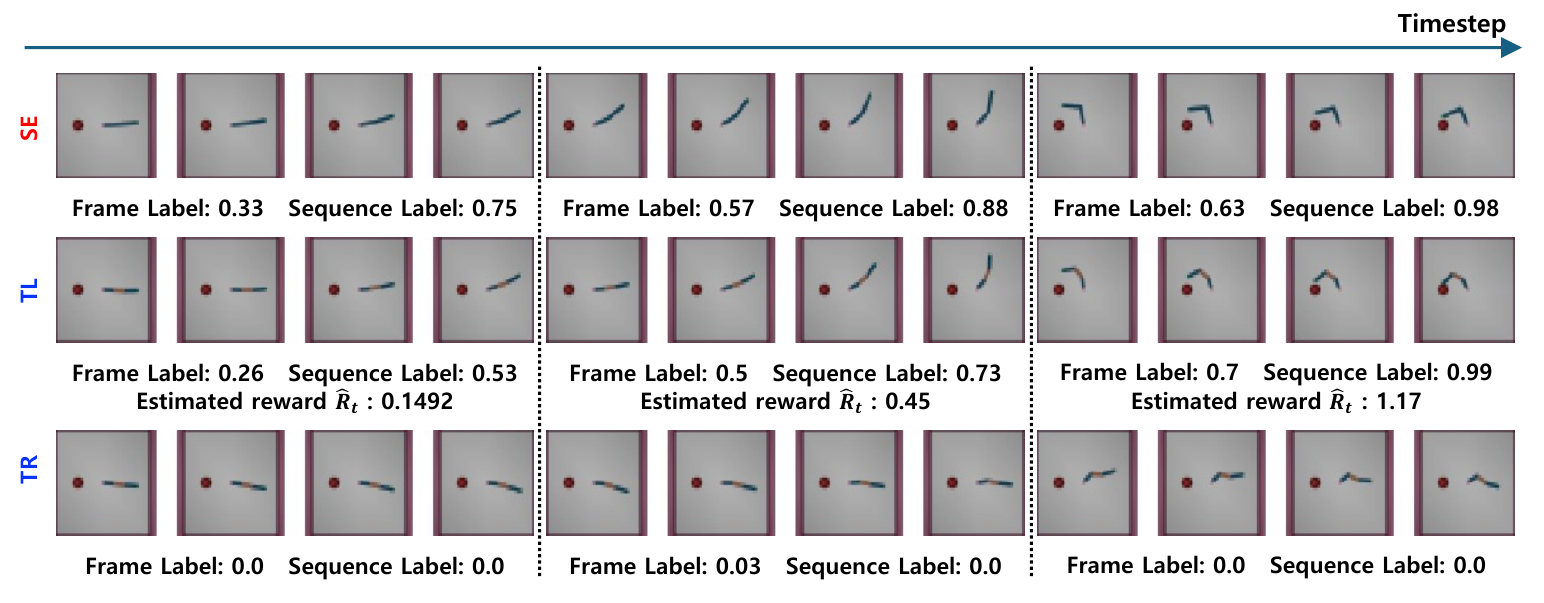}}
    \caption{Image mapping and reward analysis on RE2-to-RE3 task (Goal 1)}
    \label{app:map_re2re3}
    \centerline{\includegraphics[width=0.9\columnwidth]{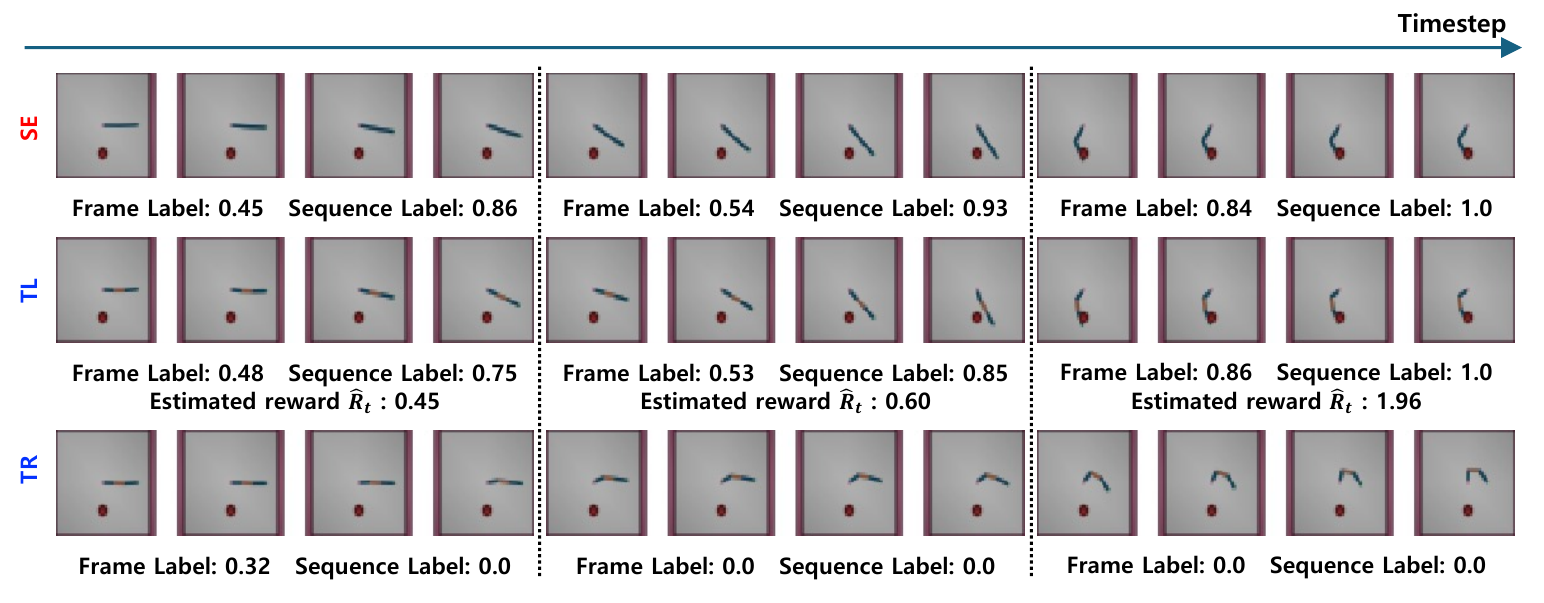}}
    \caption{Image mapping and reward analysis on RE2-to-RE3 task (Goal 2)}
    \label{app:map_re2re3_2}
    \end{center}
\end{figure}

\textbf{RE2-to-RE3 task}: 
Figures \ref{app:map_re2re3} and \ref{app:map_re2re3_2} illustrate the RE2-to-RE3 task, focusing on image mapping, reward analysis, and label estimation for two scenarios with different goals. In this task, both frame and sequence label values increase as the agent progresses toward the goal, with sequence labels nearing 1 upon reaching the target. Frame label estimates, however, vary based on the agent's proximity to the goal, as the episode continues after the goal is reached. Due to the arm's initial alignment to the right, leftward arm movements are less represented in the expert samples, resulting in higher frame label values for more common rightward movements. Similar to the IP-to-IDP task, random samples that fail to approach the goal produce sequence label values close to 0, enabling effective learning by rewarding the agent for reaching the target quickly and accurately.

\begin{figure}[!h]
    \begin{center}
    \centerline{\includegraphics[width=0.9\columnwidth]{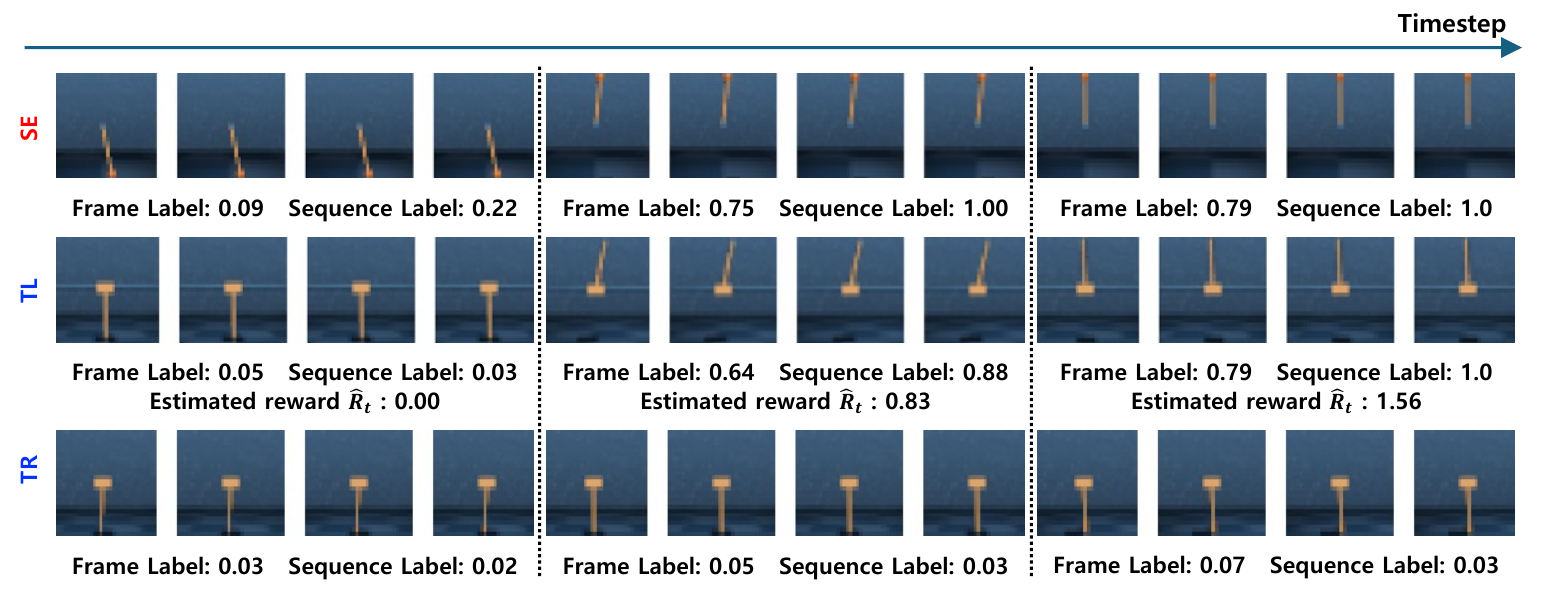}}
    \caption{Image mapping and reward analysis on Pend-to-CS task}
    \label{app:map_ptoc}
    \centerline{\includegraphics[width=0.9\columnwidth]{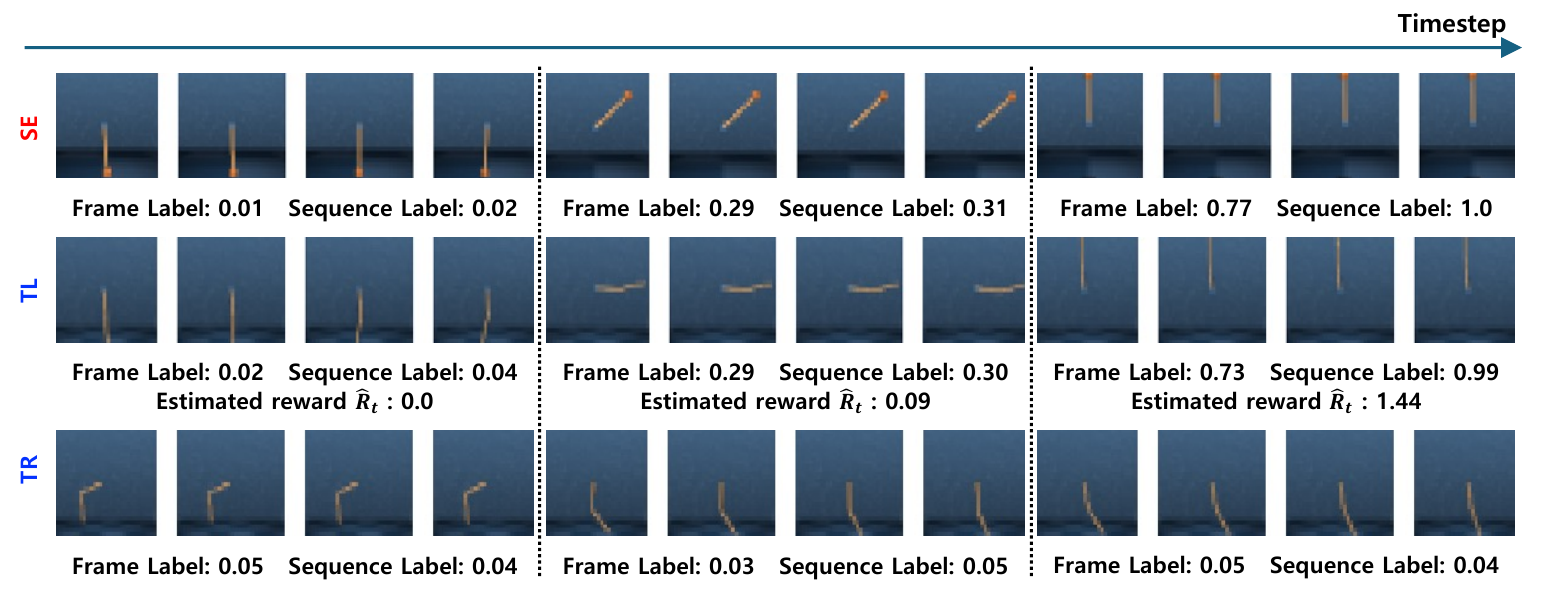}}
    \caption{Image mapping and reward analysis on Pend-to-Acrobot task}
    \label{app:map_ptoa}
    \end{center}
\end{figure}

\newpage
\textbf{DMC Pendulum task}: Figures \ref{app:map_ptoc} and \ref{app:map_ptoa} present the DMC Pend-to-CS and Pend-to-Acrobot tasks, highlighting the mapping and reward estimation in these challenging Pendulum environments. 
Initially, both frame and sequence label values are low, but they progressively increase as the agent approaches the goal, resulting in higher reward estimates. The rightmost images depict states aligned with the expert's goal, corresponding to the highest reward estimates. 
Notably, the frame label does not reach 1 even after achieving the goal. This occurs because the goal state is maintained until the episode ends, leading the frame label  to represent an average label estimation during this period.

\newpage
\subsection{MuJoCo Tasks}

\begin{figure}[!h]
    \begin{center}
    \centerline{\includegraphics[width=0.9\columnwidth]{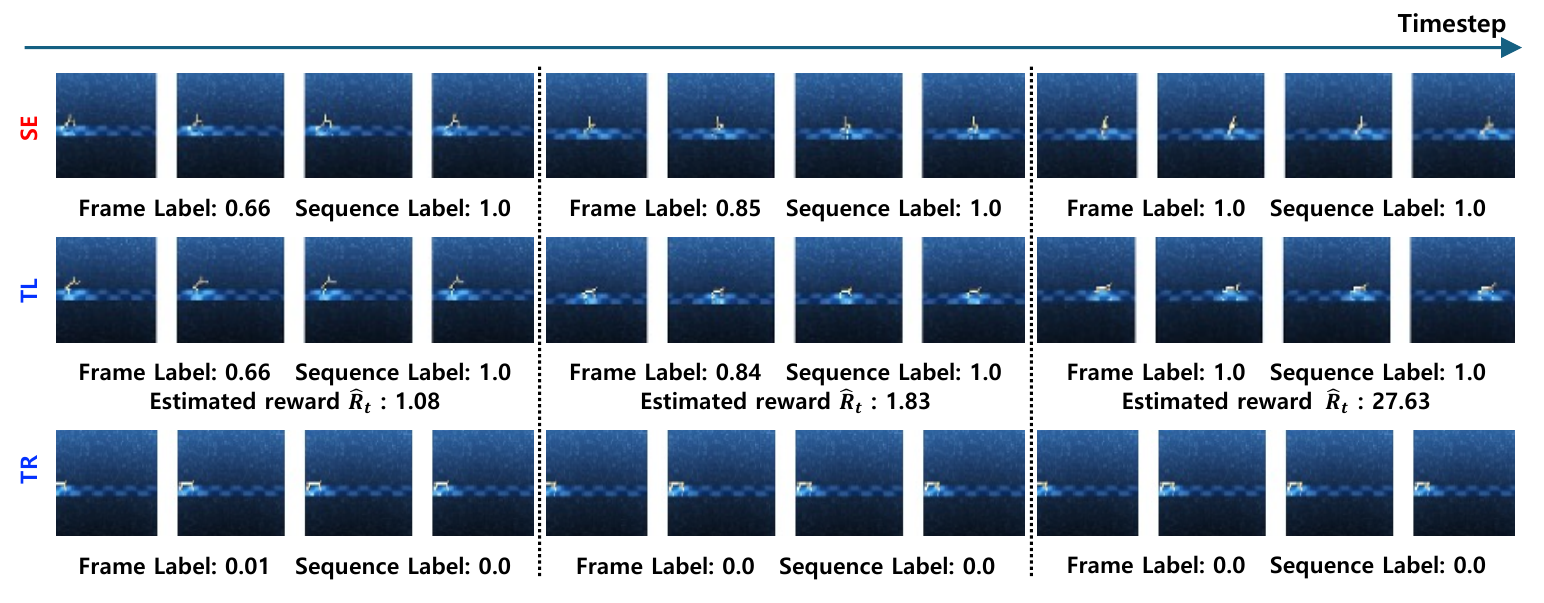}}
    \caption{Image mapping and reward analysis on Walker-to-Cheetah task}
    \label{app:map_wtoc}
    \end{center}
\end{figure}

\textbf{Walker-to-Cheetah task}: Fig. \ref{app:map_wtoc} illustrates the Walker-to-Cheetah task. At the episode's early stages, leftmost samples show label estimations and rewards assigned to observations diverging from random data. As the agent progresses, frame label predictions gradually increase, leading to higher reward estimates. By the end of the episode, both frame and sequence labels converge to values near 1, resulting in the highest reward estimates.

\begin{figure}[!h]
    \begin{center}
    \centerline{\includegraphics[width=0.9\columnwidth]{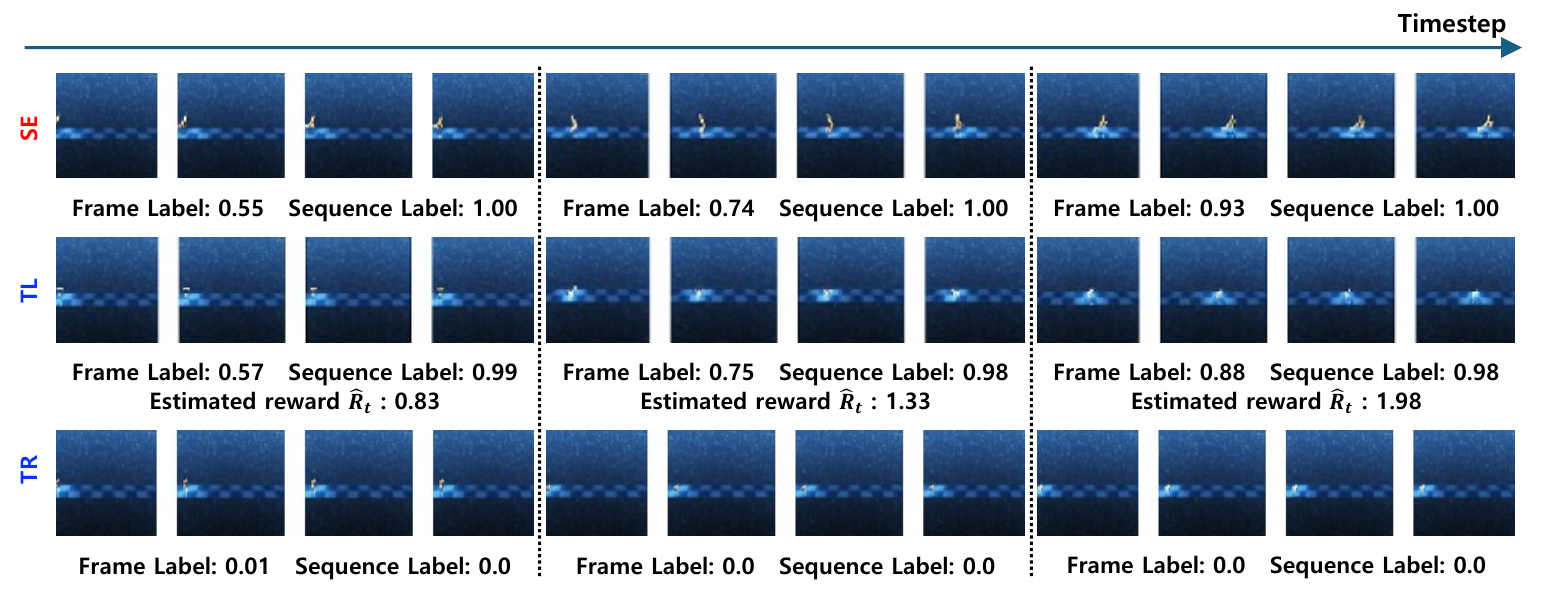}}
    \caption{Image mapping and reward analysis on Walker-to-Hopper task}
    \label{app:map_wtoh}
    \end{center}
\end{figure}

\textbf{Walker-to-Hopper task}: In Fig. \ref{app:map_wtoh}, the Walker-to-Hopper task demonstrates similar trends, with frame labels and rewards increasing as timesteps progress. Unlike conventional settings where rewards are based on maintaining torso stability, our approach rewards forward velocity, prioritizing forward movement over balance. This adjustment enables the agent to achieve positions comparable to the Hopper expert by focusing on efficient locomotion.

\begin{figure}[!h]
    \begin{center}
    \centerline{\includegraphics[width=0.9\columnwidth]{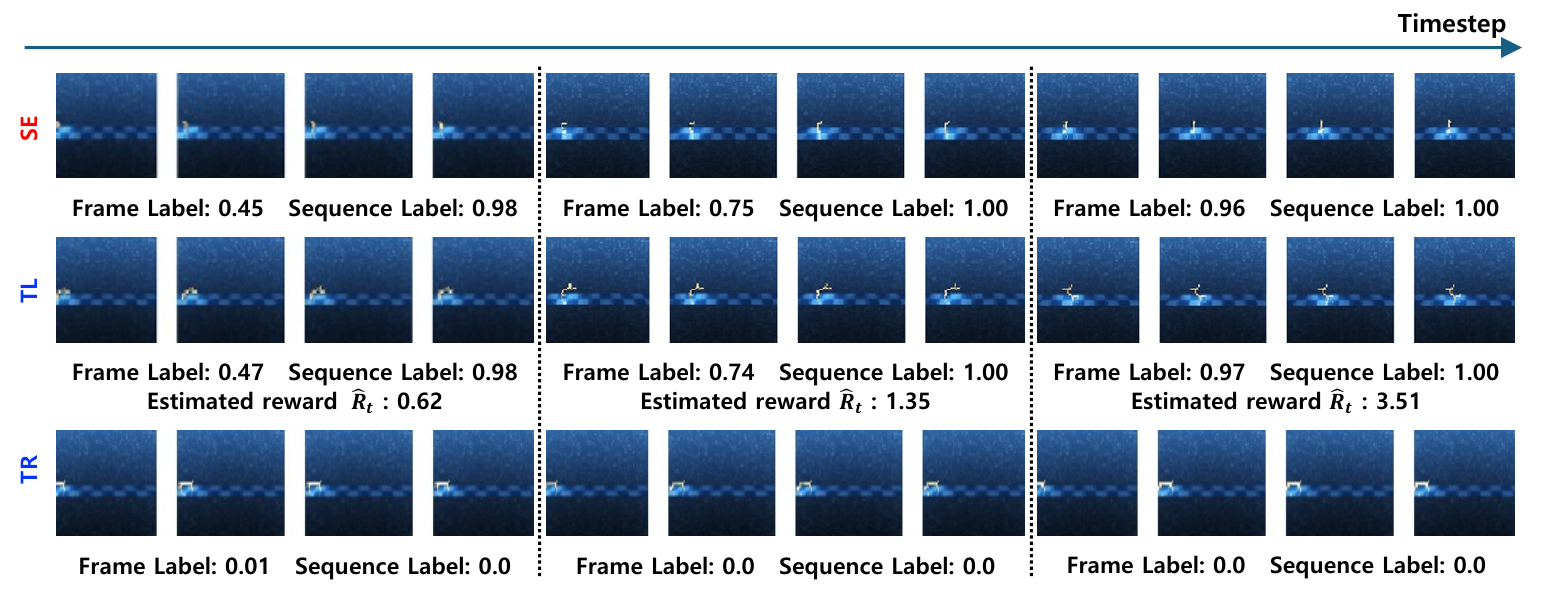}}
    \caption{Image mapping and reward analysis on Hopper-to-Cheetah task}
    \label{app:map_htoc}
    \end{center}
\end{figure}

\textbf{Hopper-to-Cheetah task}: As shown in Fig. \ref{app:map_htoc}, the frame label predictions in the Hopper-to-Cheetah task approach a value near 1 midway through the episode. This is influenced by the low velocity of the Hopper expert, which limits the forward distance achieved in the source domain. As a result, the target domain's performance is constrained by the Hopper's limitations, highlighting the challenges of domain adaptation in such cases.

\clearpage
%%%%%%%%%%%%%%%%%%%%%%%%%%%%%%%%%%%%%%%%%%%%%%%%%%%%%%%%%%%%%%%%%%%%%%%%%%%%%%%
\newpage
\section{More Ablation Studies}
\label{secapp:moreabl}

This section presents an ablation study on the discriminator coefficient $\lambda_{\textrm{disc}}$, generator coefficient 
$\lambda_{\textrm{gen}}$, and the WGAN control coefficient 
$\alpha$, which significantly influence performance but were not fully detailed in the main text. In the Pendulum tasks, we conduct parameter sweeps for IP-to-IDP, RE2-to-RE3, Pend-to-CS, and Pend-to-Acrobot. Similarly, in the MuJoCo tasks, parameter sweeps are performed for Walker-to-Cheetah and Hopper-to-Cheetah. The results of these experiments are analyzed to evaluate the impact of key parameters on performance across different task environments.

\subsection{WGAN Discrimiantor Loss Coefficient $\lambda_{\textrm{disc}}$}

The WGAN discriminator loss coefficient $\lambda_{\textrm{disc}}$ determines the degree to which source and target domain features are differentiated, with higher values increasing the feature distance between domains. For Pendulum tasks, we conducted parameter sweeps within $\lambda_{\textrm{disc}} \in [1,50]$, while for MuJoCo tasks, we used $\lambda_{\textrm{disc}} \in [0.05,1]$. The broader range for Pendulum tasks reflects their simpler and more visually similar nature compared to MuJoCo tasks, requiring a stronger discriminator to effectively separate features.

\begin{figure}[!h]
    \begin{center}
    \centerline{\includegraphics[width=\columnwidth]{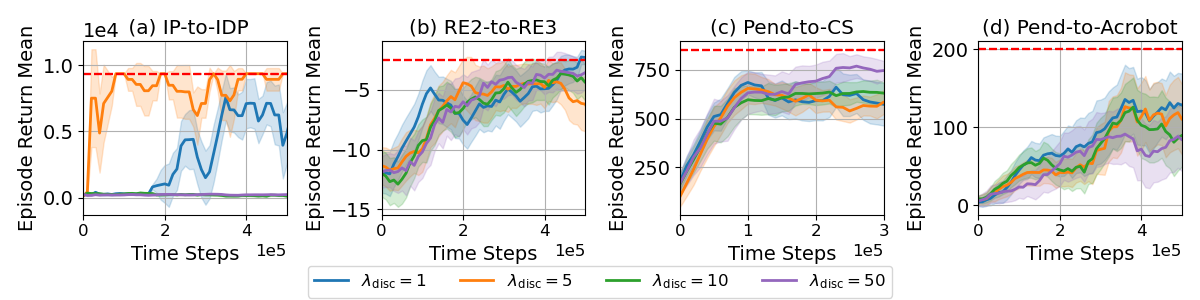}}
    \caption{Impact of the WGAN discriminator loss coefficient $\lambda_{\textrm{disc}}$ on Pendulum tasks}
    \label{app:abl_disc_loss_pend}
    \centerline{\includegraphics[width=0.8\columnwidth]{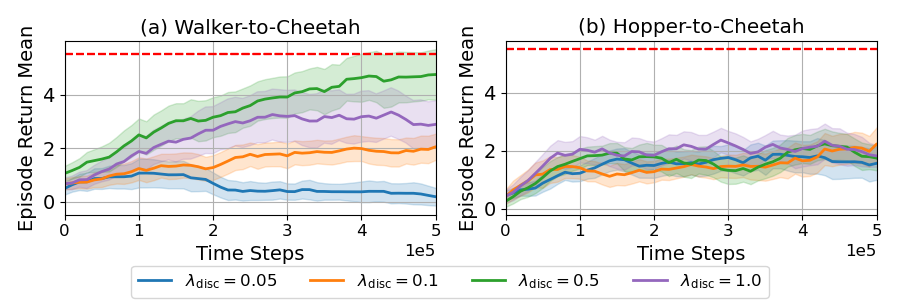}}
    \caption{Impact of the WGAN discriminator loss coefficient $\lambda_{\textrm{disc}}$ on MuJoCo tasks}
    \label{app:abl_disc_loss_muj}
    \end{center}
\end{figure}

From Fig. \ref{app:abl_disc_loss_pend}, which illustrates the impact of WGAN discriminator loss coefficient $\lambda_{\textrm{disc}}$, most Pendulum tasks show minimal sensitivity to $\lambda_{\textrm{disc}}$, with the exception of IP-to-IDP, where significant dependence is observed, and optimal performance occurs at $\lambda_{\textrm{disc}} = 5$. Similarly, Fig. \ref{app:abl_disc_loss_muj}, highlighting the impact of $\lambda_{\textrm{disc}}$ on MuJoCo tasks, indicates that Hopper-to-Cheetah is relatively unaffected by $\lambda_{\textrm{disc}}$, whereas Walker-to-Cheetah exhibits notable sensitivity, achieving its best performance at $\lambda_{\textrm{disc}} = 0.5$.

These findings highlight the importance of balancing feature separation in adversarial training. Excessively large or small $\lambda_{\textrm{disc}}$ values can hinder learning by either overly separating or insufficiently distinguishing features. The results emphasize the need for carefully tuning $\lambda_{\textrm{disc}}$ to ensure effective feature differentiation, particularly in tasks where the feature alignment process is more challenging.

\subsection{WGAN Generator Loss Coefficient $\lambda_{\textrm{disc}}$}

The WGAN generator loss coefficient $\lambda_{\textrm{gen}}$ plays the opposite role of the discriminator loss coefficient, training the encoder and decoder to reduce the distinction between features of different domains. A higher $\lambda_{\textrm{gen}}$ causes features to become less distinguishable, making it harder for the discriminator to differentiate them, whereas a lower $\lambda_{\textrm{gen}}$ allows features to remain more distinct. For Pendulum tasks, parameter search was conducted within a wide range of $\lambda_{\textrm{gen}} \in [0.05, 10]$, whereas for MuJoCo tasks, a narrower range of $\lambda_{\textrm{gen}} \in [0.01, 1]$ was used. This difference reflects the broader diversity of tasks in the Pendulum domain.

\begin{figure}[!h]
    \begin{center}
    \centerline{\includegraphics[width=\columnwidth]{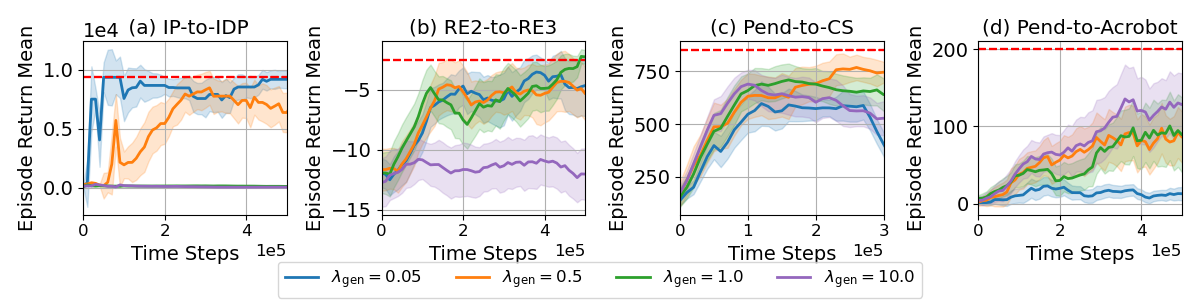}}
    \caption{Impact of the WGAN generator loss coefficient $\lambda_{\textrm{gen}}$ on Pendulum tasks}
    \label{app:abl_gen_loss_pend}
    \centerline{\includegraphics[width=0.8\columnwidth]{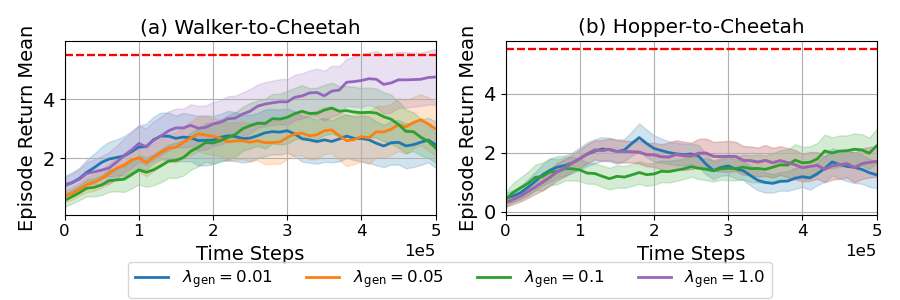}}
    \caption{Impact of the WGAN generator loss coefficient $\lambda_{\textrm{gen}}$ on MuJoCo tasks}
    \label{app:abl_gen_loss_muj}
    \end{center}
\end{figure}

Fig. \ref{app:abl_gen_loss_pend} shows the impact of $\lambda_{\textrm{gen}}$ on Pendulum tasks. While most tasks are relatively insensitive to generator loss scale, IP-to-IDP is notably affected, with performance highly dependent on the chosen value. Additionally, in RE2-to-RE3, excessively large $\lambda_{\textrm{gen}}$ values degrade performance. This is because a strong generator overly aligns features, even ignoring critical identity information required for expert distinction. For MuJoCo tasks, Fig. \ref{app:abl_gen_loss_muj} highlights the effects of $\lambda_{\textrm{gen}}$. While both environments show limited sensitivity, Walker-to-Cheetah achieves optimal performance at $\lambda_{\textrm{gen}} = 1$, with performance degrading at other scales.

In summary, similar to the discriminator loss coefficient, maintaining an appropriate balance in $\lambda_{\textrm{gen}}$ is critical. Excessive or insufficient scaling can lead to suboptimal domain adaptation, emphasizing the importance of tuning generator loss coefficients to task-specific requirements.

\newpage

\subsection{WGAN Control Coefficient $\alpha$}

\begin{figure}[!h]
    \begin{center}
    \centerline{\includegraphics[width=\columnwidth]{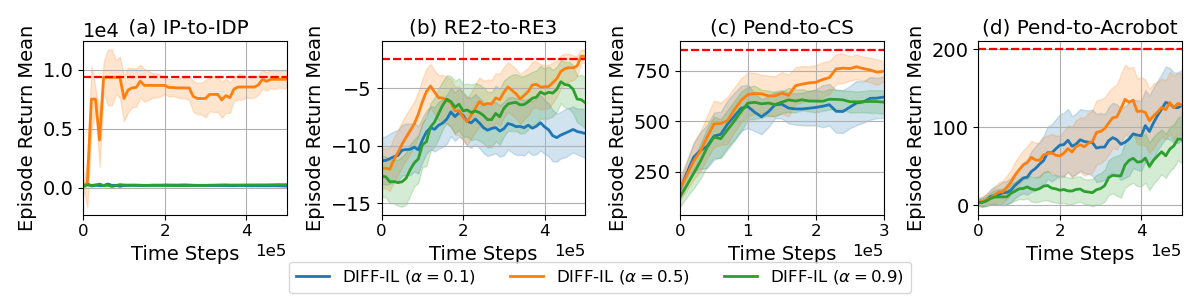}}
    \caption{Ablation study : WGAN control factor $\alpha$ on Pendulum tasks}
    \label{app:abl_alpha_pend}
    \centerline{\includegraphics[width=0.8\columnwidth]{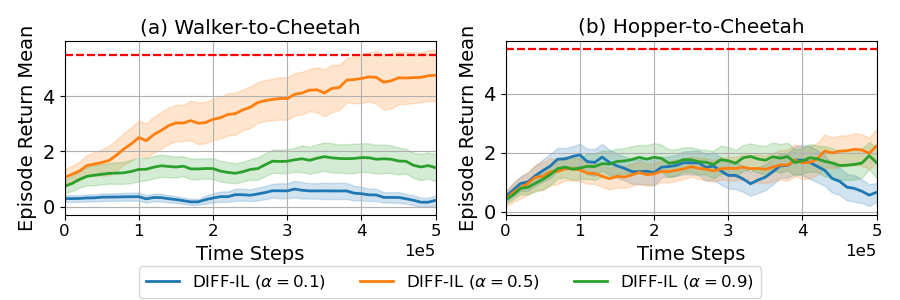}}
    \caption{Ablation study : WGAN control factor $\alpha$ on MuJoCo tasks}
    \label{app:abl_alpha_muj}
    \end{center}
\end{figure}

Lastly, an extended parameter search for the WGAN control coefficient $\alpha$, which balances frame mapping and sequence mapping, was conducted across additional tasks. For both Pendulum and MuJoCo tasks, we compared the performance at $\alpha \in [0.1, 0.5, 0.9]$. Fig. \ref{app:abl_alpha_pend} presents the results for Pendulum tasks, while Fig. \ref{app:abl_alpha_muj} shows the results for MuJoCo tasks.

As evident from these figures, $\alpha = 0.5$ consistently achieves the best performance across all considered tasks, with significant performance drops observed when either frame mapping or sequence mapping is prioritized excessively. This highlights that, in the proposed DIFF-IL, achieving domain-invariant feature extraction and effectively imitating expert behavior requires giving balanced importance to both frame mapping and sequence mapping.

\section{Limitations}

The proposed DIFF-IL effectively extracts task-relevant features through domain-invariant feature extraction and accurately estimates rewards for target learner imitation, making it a robust solution for cross-domain imitation learning. While it offers significant advantages, some limitations exist. Single-frame encoding increases memory usage, which can limit batch size and parallel execution, though this issue can be mitigated by reducing feature size or adjusting the batch size. Hyperparameter tuning, particularly for WGAN-related parameters, requires attention, but the process is simplified by the criteria provided in the ablation study, minimizing complexity. Additionally, the target learner's performance is constrained by the source expert's capabilities, reflecting an inherent limitation of imitation learning. Despite these challenges, DIFF-IL consistently demonstrates exceptional effectiveness across diverse visual environments, addressing key cross-domain imitation learning challenges. These limitations also suggest directions for future work, such as optimizing resource efficiency and overcoming performance constraints tied to the source expert. Overall, DIFF-IL's strengths far outweigh its limitations, establishing it as a powerful framework for cross-domain imitation learning.

%%%%%%%%%%%%%%%%%%%%%%%%%%%%%%%%%%%%%%%%%%%%%%%%%%%%%%%%%%%%%%%%%%%%%%%%%%%%%%%
%%%%%%%%%%%%%%%%%%%%%%%%%%%%%%%%%%%%%%%%%%%%%%%%%%%%%%%%%%%%%%%%%%%%%%%%%%%%%%%

\end{document}